\documentclass[runningheads]{llncs}
\usepackage{graphicx}
\usepackage{svg}

\usepackage[font=footnotesize]{caption}
\usepackage{subcaption}

\usepackage{array}
\usepackage{tikz}
\usepackage{amsmath,amssymb} %
\usepackage{color}

\usepackage{wrapfig}

\usepackage[accsupp]{axessibility}  %

\usepackage{hyperref}
\hypersetup{colorlinks=true, linkcolor=blue, citecolor=blue}

\usepackage{booktabs}
\usepackage{multirow}
\usepackage{textcomp}
\usepackage[capitalize]{cleveref}

\newcommand{\beginsupplement}{
        \numberwithin{figure}{section}
        \numberwithin{table}{section}
     }

\newcommand{\eat}[1]{\ignorespaces}

\newcommand{\dataset}{\texttt{MS-CXR}}
\newcommand{\cxrmodel}{CXR-BERT}
\newcommand{\jointmodel}{BioViL}
\newcommand{\localjointmodel}{BioViL-L}

\def\Plus{\texttt{+}}

\begin{document}
\pagestyle{headings}
\mainmatter
\def\ECCVSubNumber{1557}  %

\title{Making the Most of Text Semantics to Improve Biomedical Vision--Language Processing}

\titlerunning{Making the Most of Text Semantics to Improve Biomedical VLP}

\makeatletter
\def\@fnsymbol#1{\ensuremath{\ifcase#1\or *\or \dagger\or \ddagger\or
   \mathsection\or \mathparagraph\or \|\or **\or \dagger\dagger
   \or \ddagger\ddagger \else\@ctrerr\fi}}
\makeatother
\makeatletter
\newcommand{\printfnsymbol}[1]{%
  \textsuperscript{\@fnsymbol{#1}}%
}
\makeatother

\author{
Benedikt Boecking\thanks{These authors contributed equally.}\thanks{Work conducted during Benedikt Boecking's internship at Microsoft Research.} %\orcidlink{0000-0003-4822-0531}
\and Naoto Usuyama\printfnsymbol{1} %\orcidlink{0000-0003-0888-929X}
\and Shruthi Bannur % \orcidlink{0000-0001-5750-7628}
\and Daniel C. Castro\index{C. Castro, Daniel}%\orcidlink{0000-0002-6829-7045}
\and Anton Schwaighofer % \orcidlink{0000-0003-1557-0527}
\and Stephanie Hyland %\orcidlink{0000-0001-7571-6036}
\and Maria Wetscherek
\and \\Tristan Naumann %\orcidlink{0000-0003-2150-1747}
\and Aditya Nori
\and Javier Alvarez-Valle %\orcidlink{0000-0003-0906-4177}
\and \\Hoifung Poon % \orcidlink{0000-0002-9067-0918}
\and Ozan Oktay\thanks{Corresponding author: \texttt{ozan.oktay@microsoft.com}} %\orcidlink{0000-0003-2976-0874}
}

\authorrunning{B. Boecking\printfnsymbol{1}, N. Usuyama\printfnsymbol{1} et al.}

\institute{Microsoft Health Futures}

\maketitle

\begin{abstract}
Multi-modal data abounds in biomedicine, such as radiology images and reports. Interpreting this data at scale is essential for improving clinical care and accelerating clinical research. 
Biomedical text with its complex semantics poses additional challenges in vision--language modelling compared to the general domain, and previous work has used insufficiently adapted models that lack  domain-specific language understanding. 
In this paper, we show that principled textual semantic modelling can substantially improve contrastive learning in self-supervised vision--language processing. 
We release a language model that achieves state-of-the-art results in radiology natural language inference through its improved vocabulary and novel language pretraining objective leveraging semantics and discourse characteristics in radiology reports. 
Further, we propose a self-supervised joint vision--language approach with a focus on better text modelling. It establishes new state of the art results on a wide range of publicly available benchmarks, in part by leveraging our new domain-specific language model. 
We release a new dataset with locally-aligned phrase grounding annotations by radiologists to facilitate the study of complex semantic modelling in biomedical vision--language processing. A broad evaluation, including on this new dataset, shows that our contrastive learning approach, aided by textual-semantic modelling, outperforms prior methods in segmentation tasks, despite only using a global-alignment objective.
\end{abstract}

\section{Introduction}
Advances in deep learning have enabled automated diagnosis systems that operate near or above expert-level performance, paving the way for the use of machine learning systems to improve healthcare workflows, for example by supporting fast triaging and assisting medical professionals to reduce errors and omissions~\cite{chilamkurthy2018deep,esteva2017dermatologist,miura2021improving,titano2018automated}. 
A major hurdle to the widespread development of these systems is a requirement for large amounts of detailed ground-truth clinical annotations for supervised training, which are expensive and time-consuming to obtain.
Motivated by this challenge, there has been a rising interest in multi-modal self-supervised learning~\cite{huang2021gloria,liao2021multimodal} and 
cross-modal weak supervision~\cite{dunnmon2020cross,eyuboglu2021multi,irvin2019chexpert,titano2018automated,wang2017chestx} (using partial and imperfect labels derived from the auxiliary modality), in particular for paired image--text data. Such data is collected routinely during clinical practice, and common examples are X-ray images~\cite{dunnmon2020cross,irvin2019chexpert,wang2017chestx} or computed tomography (CT) scans~\cite{chilamkurthy2018deep,dunnmon2020cross,eyuboglu2021multi,titano2018automated} paired with reports written by medical experts.
Importantly, while many remain private, some paired clinical datasets~\cite{bustos2020padchest,demner2016preparing,johnson2019mimic} have been released to the research community such as MIMIC-CXR~\cite{johnson2019mimic}.

This article focuses on self-supervised vision--language processing (VLP) for paired image and text data in the biomedical domain. The goal is to jointly learn good image and text representations that can be leveraged by downstream applications such as zero-/few-shot image classification, report generation and error detection, and disease localisation.
Self-supervised VLP has several advantages over supervised learning, not just because it does not require laborious manual annotations, but also because it does not operate on a fixed number of predetermined conditions or object categories, since the joint latent space is learned from raw text. However, in contrast to the general domain setting, self-supervised VLP with biomedical data poses additional challenges. Take radiology as an example, publicly available datasets~\cite{johnson2019mimic,demner2016preparing,bustos2020padchest} are usually smaller, on the order of a few hundred thousand pairs rather than millions in general-domain vision--language processing (e.g.~\cite{radford2021learning} collected 400M text--image pairs on the Internet for self-supervision). Furthermore, linguistic challenges are different in biomedical settings, including common usage of negations, expressions of uncertainty, long-range dependencies, more frequent spatial relations, the use of domain-specific modifiers, as well as scientific terminology rarely found in the general domain. Taking negation as an example, ``there is no dog in this picture'' would be a highly unusual caption on social media, but ``there is no evidence of pneumonia in the left lung'' or ``there are no new areas of consolidation to suggest the presence of pneumonia'' are descriptions commonly found in radiology reports.
Moreover, pretrained models including object detectors often used in general domain visual grounding are typically unavailable or under-perform in domain-specific applications (see also Supp.\ in \cite{huang2021gloria}). Additionally, imbalance in underlying latent entities of interest (e.g., pulmonary findings) can cause larger numbers of false negatives in contrastive learning objectives that sample at random, which can lead models to degrade and memorise irrelevant text and image aspects. For example, radiology images and text reports with normal findings occur much more frequently compared to exams that reveal abnormal conditions such as pneumonia or pneumothorax (also see~\cite{dai2021bdkg}). Supp.~\ref{sec:background} provides further discussion of these challenges. 

Related self-supervised VLP work \cite{hsu2018unsupervised,huang2021gloria,liao2021multimodal,muller2021joint,zhang2020contrastive} has achieved impressive downstream classification and zero-shot classification performance. However, our study reveals that suboptimal text modelling due to insufficient vocabulary adjustment, fine-tuning, and language grounding appears to have gone unnoticed, all of which are shown to degrade the quality of joint latent representations. In particular, a more thorough benchmarking of the text, image, and shared embeddings, across a multitude of downstream benchmarks, reveals that large improvements in performance are possible by taking care to build highly specialised text models and by maintaining their performance during joint training. Free-text image descriptions provide a semantically dense learning signal compared to image-only contrastive methods and supervised classification \cite{desai2021virtex}. Further, extracting shared semantics of images and text pairs is easier for text, as the modality is already discretised. Thus, making the most of text modelling before and during joint training can lead to large improvements in not just the text model, but also of the image model and joint representations. 
\begin{figure}[t]
\centering{\includegraphics[width=\textwidth,height=\textheight,keepaspectratio,trim={1pt 2pt 0 6pt},clip]{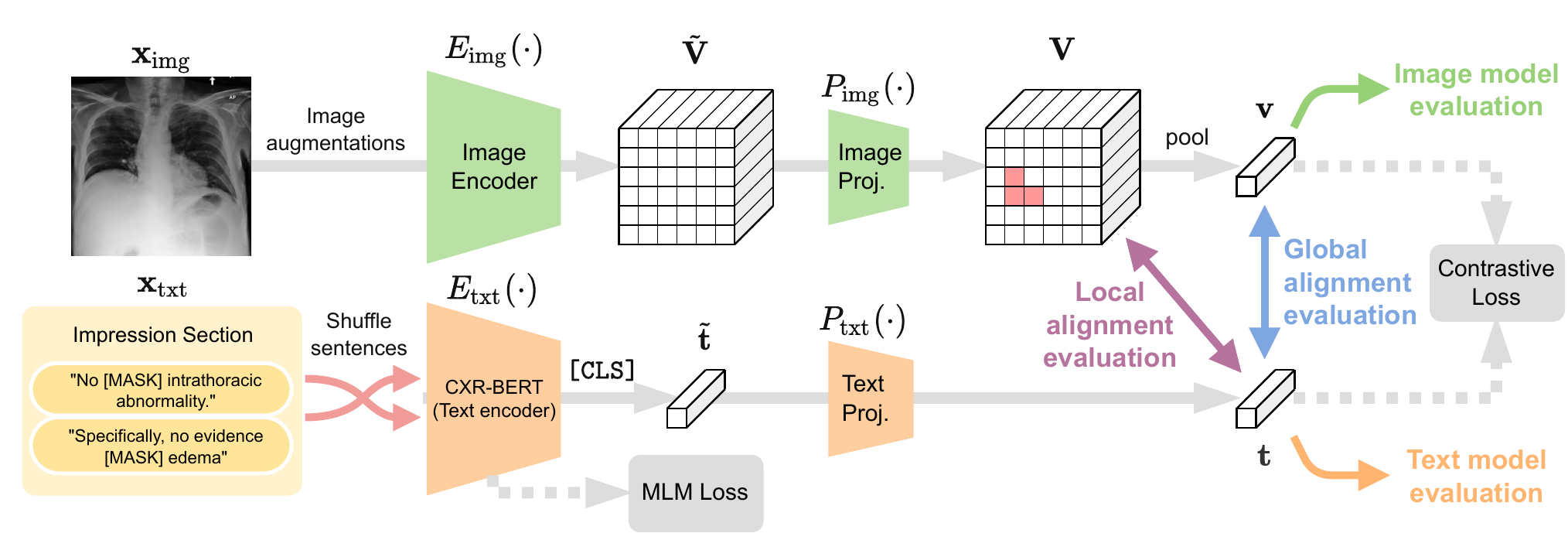}}
\caption{\jointmodel\ leverages our radiology-specific text encoder (\cxrmodel), text augmentation, regularisation, and maintains language model quality via a masked language modelling (MLM) loss. We conduct a broad evaluation of models and representations that includes zero-shot classification, phrase grounding, and natural language inference. \label{fig:mainssvlp}}\vspace{-20pt}
\end{figure}
We present the following contributions in this work:
\begin{enumerate}
    \item We introduce and release a new chest X-ray (CXR) domain-specific language model, \cxrmodel\footnote{Pretrained models available on HuggingFace: \url{https://aka.ms/biovil-models}} (\cref{fig:cxrbert}). Through an improved vocabulary, a novel pretraining procedure, regularisation, and text augmentation, the model considerably improves radiology natural language inference~\cite{miura2021improving}, radiology masked token prediction~\cite{devlin2018bert,liu2019roberta}, and downstream VLP task performance.
    \item We propose and release a simple but effective self-supervised VLP approach for paired biomedical data, which we name \jointmodel\footnotemark[\value{footnote}]\footnote{Code can be found at:~\url{https://aka.ms/biovil-code}}\ (\cref{fig:mainssvlp}), and evaluate in the radiology setting.
    Through improvements in text modelling, text model grounding, augmentation, and regularisation, the approach yields new state-of-the-art performance on a wide range of public downstream benchmarks. Our large-scale evaluation (see \cref{table:evaluation_types}) includes phrase grounding, natural  language inference~\cite{miura2021improving}, as well as
    zero-/few-shot classification and zero-shot segmentation via the RSNA Pneumonia dataset~\cite{shih2019augmenting,wang2017chestx}. Notably, our approach achieves improved segmentation performance despite only using a global alignment objective during training.
    \item We also release a \textit{\textbf{L}ocal \textbf{A}lignment \textbf{Che}st \textbf{X}-ray dataset}, \dataset\footnote{The \dataset\ dataset can be found on PhysioNet \url{https://aka.ms/ms-cxr}.}, to encourage reproducible evaluation of shared latent semantics learned by biomedical image-text models. 
    This large, well-balanced phrase grounding benchmark dataset contains carefully curated image regions annotated with descriptions of eight radiology findings, as verified by board-certified radiologists.
    Unlike existing chest X-ray benchmarks, this challenging phrase grounding task evaluates joint, local image-text reasoning while requiring real-world language understanding, e.g.\ to parse domain-specific location references, complex negations, and bias in reporting style.
\end{enumerate}

\vspace{-12pt}
\section{Making the Most of Free-Text Supervision}
\vspace{-5pt}
\newcommand{\RSMloss}{\mathcal{L}_\mathrm{RSM}}
\newcommand{\MLMloss}{\mathcal{L}_\mathrm{MLM}}
\newcommand{\MLMweight}{\lambda_\mathrm{MLM}}
\newcommand{\GAloss}{\mathcal{L}_\mathrm{GA}}
\newcommand{\GAweight}{\lambda_\mathrm{GA}}
\newcommand{\jointloss}{\mathcal{L}_\mathrm{joint}}
\newcommand{\CSEloss}{\mathcal{L}_\mathrm{CSE}}
\newcommand{\setcls}{\mathcal{S}}
\newcommand{\precls}{\tilde{\mathbf{t}}} %
\newcommand{\preclsimp}{\precls^\mathrm{I}} %
\newcommand{\preclsfin}{\precls^\mathrm{F}} %
\newcommand{\cls}{\mathbf{t}}
\newcommand{\clsimp}{\cls^\mathrm{I}}
\newcommand{\clsfin}{\cls^\mathrm{F}}
\newcommand{\subimg}{img}
\newcommand{\subtxt}{txt}
\newcommand{\inputimg}{\mathbf{x}_\mathrm{\subimg}}
\newcommand{\inputtxt}{\mathbf{x}_\mathrm{\subtxt}}
\newcommand{\phrase}{\mathbf{w}}
\newcommand{\corpus}{\mathcal{D}}
\newcommand{\batch}{\mathcal{B}}

\newcommand{\encimg}{E_\mathrm{\subimg}}
\newcommand{\enctxt}{E_\mathrm{\subtxt}}
\newcommand{\baseproj}{P}
\newcommand{\projimg}{\baseproj_\mathrm{\subimg}}
\newcommand{\projtxt}{\baseproj_\mathrm{\subtxt}}
\newcommand{\localrep}{\mathbf{V}}
\newcommand{\localemb}{\tilde{\mathbf{V}}}
\newcommand{\globalrep}{\mathbf{v}}
\newcommand{\probeweights}{\boldsymbol{\beta}}
\newcommand{\clstoken}{\texttt{[CLS]}}
\newcommand{\Findings}{\textsc{Findings}}
\newcommand{\Impression}{\textsc{Impression}}

We assume that we are given a set $\corpus$ of pairs of radiology images and reports $(\inputimg,\inputtxt)$.  Let $\phrase=(w_1,\dots,w_T)$ denote a vector of $T$ (sub-)word tokens of a text document $\inputtxt$ (after tokenisation). Recall that a BERT \cite{vaswani2017attention} encoder $\enctxt$ outputs a feature vector for each input token $w_t$ as well as a special global \clstoken\ token used for downstream classification. Let $\precls=[\enctxt(\phrase)]_\text{\clstoken}$ denote the  \clstoken\ token prediction by $\enctxt$ based on input $\phrase$, 
and $\cls=\projtxt(\precls)$ its lower-dimensional projection by a model $\projtxt$. 
\vspace{-5pt}
\subsection{\cxrmodel: Domain-Specific Language Model Pretraining}\label{sec:cxrmodel}
\begin{figure}[t]
\centering{\includegraphics[width=\textwidth,height=\textheight,keepaspectratio]{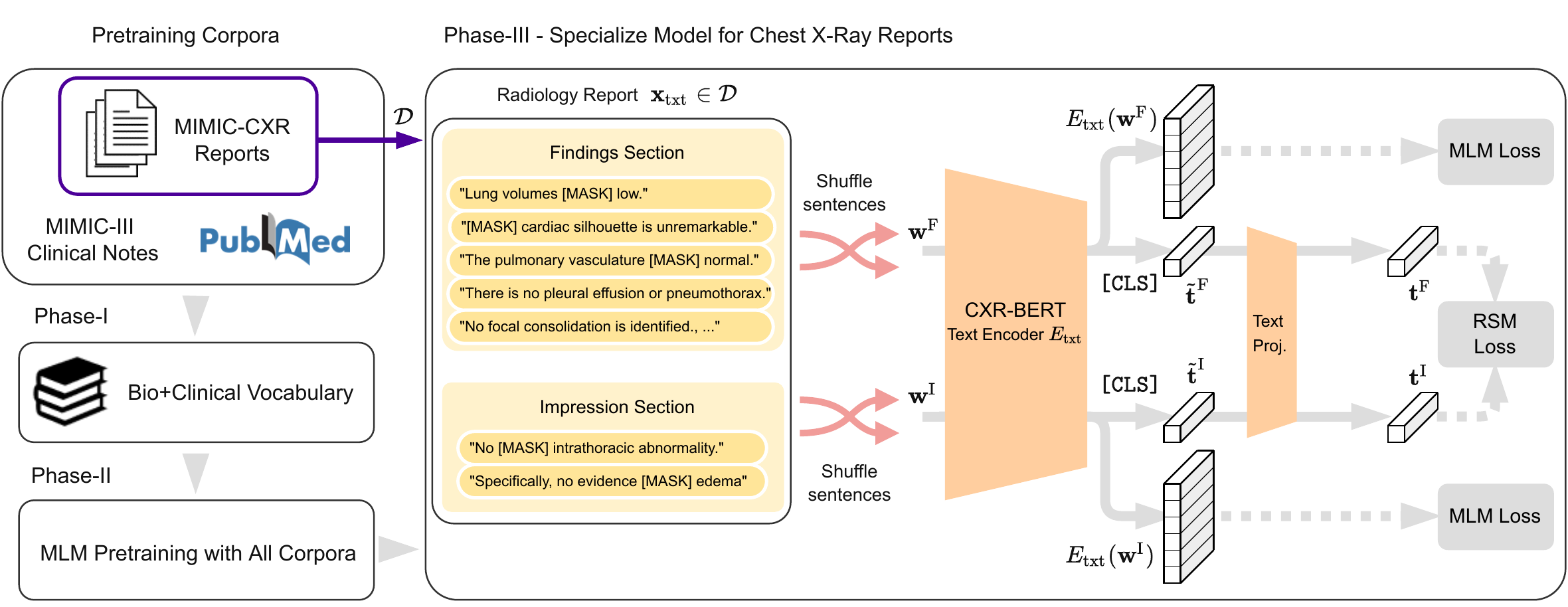}}
\caption{The proposed \cxrmodel\ text encoder has three phases of pretraining and uses a domain-specific vocabulary, masked language modelling (MLM) and radiology section matching (RSM) losses, regularisation, and text augmentations.  \label{fig:cxrbert}}\vspace*{-10pt}
\end{figure}

\begin{wraptable}{r}{6.8cm}
\vspace{-24pt}
\centering
\caption{Vocabulary
comparison of common radiology terms
with ClinicalBERT (Wiki/Book, cased), PubMedBERT (PubMed, uncased), and \cxrmodel\ (PubMed\Plus MIMIC-III/CXR, uncased). $\checkmark$~marks that a word appears in the vocabulary, otherwise its sub-tokens are shown.}
\label{tab:vocab}
\setlength{\tabcolsep}{.5em}
\resizebox{\linewidth}{!}{
\begin{tabular}{@{}llll@{}}
\toprule
       Full word & ClinicalBERT & PubMedBERT & \cxrmodel\\
\midrule
       pneumonia   & \checkmark & \checkmark & \checkmark \\
       opacity & op-acity & \checkmark & \checkmark \\
       effusion    &     e-ff-usion    & \checkmark & \checkmark \\
       pneumothorax&     p-ne-um-oth-orax & \checkmark & \checkmark \\
       atelectasis &     ate-lect-asis &  ate-le-ct-asis  &     \checkmark  \\
       cardiomegaly&     card-io-me-gal-y & cardio-me-gal-y & \checkmark \\
       bibasilar & bi-bas-ila-r & bib-asi-la-r & \checkmark \\
\bottomrule
\end{tabular}
}{}
\vspace{-20pt}
\end{wraptable}
We introduce \cxrmodel\ (\cref{fig:cxrbert}), a specialised chest X-ray (CXR) language model with an adjusted vocabulary, pretrained in three phases to capture dense semantics in radiology reports \cite{casey2021systematic}. To achieve this specialisation to the CXR report domain despite limited data availability, our approach includes pretraining on larger data from closely related domains. The phases proceed as follows: 
\textbf{(I)}~First, we construct a custom WordPiece~\cite{wu2016google} vocabulary of 30k tokens from PubMed abstracts\footnote{Obtained via \url{https://pubmed.ncbi.nlm.nih.gov/download/}} (15\,GB), MIMIC-III~\cite{johnson2016mimic} clinical notes (3.5\,GB), and MIMIC-CXR radiology reports (0.1\,GB). With this custom vocabulary, our model produces fewer sub-word breakdowns (\cref{tab:vocab}).
\textbf{(II)}~Second, we pretrain a randomly initialised BERT model via Masked Language Modelling (MLM) on the PubMed + MIMIC-III + MIMIC-CXR corpora.
We largely follow RoBERTa~\cite{liu2019roberta} pretraining configurations, i.e.\ dynamic whole-word masking for MLM and packing of multiple sentences into one input sequence. %
This phase aims to build an initial domain-specific BERT model in the biomedical and clinical domains.
\textbf{(III)}~Third, we continue pretraining on MIMIC-CXR only to further specialise our \cxrmodel\ to the CXR domain. Here, we also add a novel sequence prediction task to the objective to obtain better sequence representations, as explained below.

Note that a raw radiology report $\inputtxt$ typically consists of several sections, including a `\Findings' section that details  clinical observations, and an `\Impression' section summarising the clinical assessment~\cite{wallis2011radiology,wilcox2006written}. Our sequence prediction objective of phase (III) aims to take advantage of this structure. Specifically, we continually run MLM pretraining on MIMIC-CXR radiology reports and propose to add a radiology section matching (RSM) pretraining task, formulated to match \Impression\ to \Findings\ sections of the same study.

Let $\theta$ denote the weights of our language model and $m \subset \{1,\dots,T\}$ denote mask indices for $M$ masked tokens, randomly sampled for each token vector $\phrase$ at every iteration. Given a batch $\batch$ of token vectors $\phrase=(w_1,\dots,w_T)$, we write the MLM loss as the cross-entropy for predicting the dynamically masked tokens: $\MLMloss = -\frac{1}{|\batch|} \sum_{\phrase \in \batch} \log p_\theta(\phrase_m \,|\, \phrase_{\backslash m}) \,.$ Further, let $(\preclsfin_i,\preclsimp_i)$ denote a pair of \clstoken\ tokens corresponding to the \Findings\ and \Impression\ sections of the same $i$\textsuperscript{th} report, and let $(\clsfin_i,\clsimp_i)$ denote the pair projected to a lower dimension via a two-layer perceptron $\projtxt$. We introduce a contrastive loss on the text modality that favours \Impression\ and \Findings\ text pair from the same report over unmatched ones. Specifically, for a batch of $N$ such pairs, the RSM loss is defined as:
\begin{equation}
    \RSMloss = -\frac{1}{N} \sum_{i=1}^N \left( \log \frac{\exp( \clsfin_i\cdot\clsimp_i /\tau_1) }{\sum_{j=1}^N \exp( \clsfin_i\cdot\clsimp_j /\tau_1) } + \log \frac{\exp( \clsimp_i\cdot\clsfin_i /\tau_1) }{\sum_{j=1}^N \exp( \clsimp_i\cdot\clsfin_j /\tau_1) } \right),
\end{equation}
where $\tau_1 > 0$ is a scaling parameter to control the margin.
The resulting total loss of the specialisation phase (III) is
$
  \mathcal{L}_\mathrm{III} = \RSMloss+ \MLMweight\MLMloss  
$.
An additional important component for regularising the RSM loss is the use of increased dropout (25\%), including on attention. We set $\tau_1 = 0.5$ and $\MLMweight = 0.1$, determined by a limited grid-search measuring $\GAloss$ (\cref{eq:global_alignment}) of the joint model on a validation set. We also note that similar losses to the RSM loss, over the same or separate text segments, have been explored successfully for sentence representation learning~\cite{gao2021simcse,logeswaran2018efficient} in other settings. As such, we empirically observed that an objective as in \cite{gao2021simcse} using masked \Findings\ to \Findings\ matching can achieve similar performance and may be an appropriate replacement in other biomedical settings with differing text structure.

\vspace{-5pt}
\paragraph{Text Augmentation.}
As domain-specific datasets are often quite small, effective text augmentation can induce large benefits.
In the radiology domain, the sentences of the \Findings\ and \Impression\ sections, which contain the detailed description and summary of the radiological findings, are usually permutation-invariant on the sentence level (cf.~\cite{preechakul2021set}).
We thus find that randomly shuffling sentences within each section is an effective text-augmentation strategy for both pretraining of \cxrmodel\ as well as during joint model training. 

\subsection{\jointmodel: Vision-Language Representation Learning}\label{sec:jointmodel}

We now introduce \jointmodel, a simple but effective self-supervised VLP setup for the biomedical domain (\cref{fig:mainssvlp}), which we study in a chest X-ray (CXR) application setting. \jointmodel uses a convolutional neural network (CNN)~\cite{lecun1989backpropagation} image encoder $\encimg$, our \cxrmodel\ text encoder $\enctxt$, and projection models $\projimg$ and $\projtxt$ to learn representations in a joint space.
The CNN model allows us to obtain a grid of local image embeddings $\localemb = \encimg(\inputimg)$, which is fine-grained enough to be useful for segmentation (e.g.\ 16$\times$16). Each encoder is followed by a modality-specific two-layer perceptron projection model $\baseproj$, which projects the encoded modality to a joint space of 128 dimensions--e.g.,  $\localrep = \projimg(\localemb)$--where the representation is $\ell_2$-normalised.
Note that projection should be applied to local embeddings before mean-pooling $\globalrep=\operatorname{pool}(\projimg(\localemb))$, which gives us the global image embedding $\globalrep$.
The text branch uses the \Impression\ section's projected \clstoken\ token $\clsimp$ as the text representation in the joint space, as it contains a succinct summary of radiological findings. 
To align the representations and learn a joint embedding, we propose to use two loss terms. For a batch of size $N$, a symmetric contrastive loss~\cite{oord2018representation} for \textit{global alignment} of the image and text projections helps us learn the shared latent semantics:
\begin{equation}\label{eq:global_alignment}
    \GAloss = -\frac{1}{N}\sum_{i=1}^N \left( \log \frac{\exp( \globalrep_i\cdot\clsimp_i  / \tau_2) }{\sum_{j=1}^N \exp( \globalrep_i\cdot\clsimp_j  / \tau_2) } + \log \frac{\exp( \clsimp_i\cdot\globalrep_i  / \tau_2) }{\sum_{j=1}^N \exp( \clsimp_i\cdot\globalrep_j  / \tau_2) } \right).
\end{equation}
where $\tau_2>0$ is a scaling parameter.
Further, we maintain the $\MLMloss$ loss during joint training, resulting in the final joint loss 
$
    \jointloss=\GAweight \GAloss+\MLMloss
$.
We set $\tau_2=0.5$ and $\GAweight = 0.5$, determined by a limited grid search measuring $\GAloss$ on a validation set.

\vspace{-5pt}
\paragraph{Augmentations, Regularisation, and Image Encoder Pretraining.}
Due to the small dataset sizes expected in biomedical applications, we use image and text augmentations to help learn known invariances.
We use a ResNet-50~\cite{he2016deep} architecture as our image encoder and pretrain the model on MIMIC-CXR images using SimCLR \cite{sicmlr} with domain-specific augmentations as detailed in \cref{sec:experiments_setup}.
For text, we use the same sentence-shuffling augmentation as in pretraining of \cxrmodel\ (see \cref{sec:experiments_setup} for details).
Furthermore, as in phase (III) of \cxrmodel\ training, we apply higher text encoder dropout (25\%) than in standard BERT settings \cite{devlin2018bert,vaswani2017attention}.
We find that the combination of all these components, including continuous MLM optimisation, is important to improve downstream performance across the board (see ablation in \cref{table:ablation_study}).

\vspace{-5pt}
\paragraph{Zero-shot Classification.}
After joint training, we use text prompts to cast the zero-shot classification problem into an image--text similarity task as in \cite{huang2021gloria,radford2021learning,rao2021denseclip}. For $C$ classes, subject-matter experts design $C$ text prompts representing the target labels $c \in \{1, \dots, C \}$, e.g.\ for presence or absence of pneumonia (see \cref{sec:rsnalocalisation}). Each class prompt is represented as a vector of tokens $\phrase^c$ and passed to the text encoder and projector of \jointmodel\ to obtain $\ell_2$-normalised text features $\cls^c=\projtxt(\enctxt(\phrase^c)) \in \mathbb{R}^{128}$.  
For each input image $\inputimg \in \mathbb{R}^{H \times W}$, we use the image encoder and projection module to obtain patch embeddings $\localrep = \projimg(\encimg(\inputimg))\in \mathbb{R}^{\frac{H}{16} \times \frac{W}{16} \times 128}$ for segmentation tasks or the pooled embedding $\globalrep=\operatorname{pool}(\localrep) \in \mathbb{R}^{128}$ for instance-classification. We use dilated convolutions \cite{yu2015multi} to obtain higher-resolution feature maps. Probabilities for classes/regions can then be computed via a softmax over the cosine similarities between the image (or region) and prompt representations.
\vspace{-5pt}
\paragraph{Few-shot Tasks with \jointmodel.} To further assess the representation quality, linear probing is applied to local ($\localrep$) and global ($\globalrep$) image representations, by learning $\probeweights \in \mathbb{R}^{128 \times C}$ weights and a bias term. Unlike \cite{huang2021gloria,zhang2020contrastive}, we leverage the pretrained projectors and class text embedding $\cls^c$ from the zero-shot setting by using them for initialisation, which leads to improved performance and further reduces the need for manual label collection. Specifically, in few-shot classification settings, the weights and bias are initialised with $\probeweights = [\cls^1, \dots, \cls^C]$ and zeros, respectively.

\vspace{-5pt}
\section{Evaluating Self-Supervised Biomedical VLP}
Accurate local alignment between modalities is an important characteristic of successful joint image-text training in healthcare, in particular since image and report samples often contain multiple clinical findings, each of which correspond to distinct image regions. Standard global-alignment approaches may attain high classification accuracy by overfitting to spurious image features for a given finding (e.g., chest tubes in images correlating with mentions of pneumothorax in reports). 
Image classification, the most frequently evaluated downstream task in related work \cite{huang2021gloria,liao2021multimodal,muller2021joint,zhang2020contrastive}, requires only scene-level labels, hence a less sophisticated understanding of natural-language image descriptions. Image classification tasks can largely be solved by simply detecting a small set of words and maintaining some understanding of negation, as exemplified by the development of automated, rule-based text-labellers such as CheXpert~\cite{irvin2019chexpert}. 
Instance-level image-text retrieval tasks address some evaluation limitations, but do not require the level of language reasoning needed to solve local correspondence between phrases and image regions.
Existing public CXR benchmark datasets to evaluate local aspects of VLP have one or more of the following limitations (see \cref{sec:relatedwork} and Supp. \ref{appendix:ourdataset},\ref{sec:app_relatedwork} for more details): bounding boxes without corresponding free text descriptions, a limited number of samples, a limited number of abnormalities, and non-curated phrases impacting evaluation quality.

With this motivation in mind, we design \dataset, a radiology visual-grounding benchmark that has domain-specific language (e.g., paraphrasing and negations) and forms a more  challenging real-world image-text reasoning task compared to existing evaluation datasets. To name just a few challenges, the phrase grounding task requires the ability to parse domain specific location modifiers, the ability to deal with reporting style biases, and understanding of complex negations, all while relating the correct findings to specific image regions. 

\begin{figure}[t]
\centering{
\includegraphics[width=\textwidth,height=\textheight,keepaspectratio]{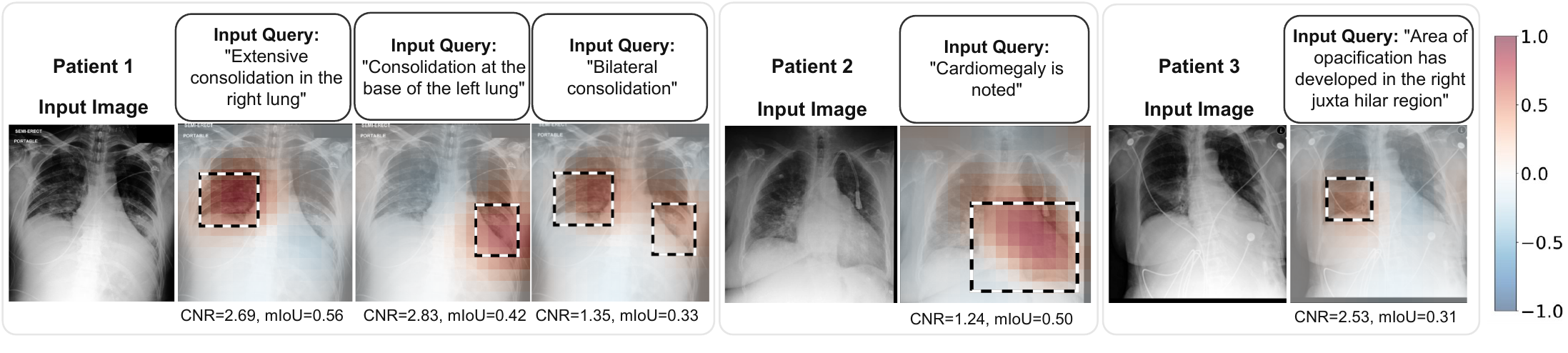}
}
\vspace*{-15pt}
\caption{Examples from the newly released \dataset ~phrase grounding dataset with \jointmodel\ latent vector similarity for different input text queries superimposed as heatmaps. Dashed boxes are ground-truth annotations by radiologists. X-ray images are mirrored horizontally.}
\label{fig:grounding_examples}
\vspace*{-15pt}
\end{figure}
\vspace{-6pt}

\subsection{MS-CXR -- A Chest X-ray Phrase Grounding Benchmark}\label{sec:ourdataset}
We publicly release \dataset, a new dataset containing image bounding box labels paired with radiology text descriptions, annotated and verified by two board-certified radiologists (see examples in \cref{fig:grounding_examples,fig:datasetexamples}). \dataset\  provides 1153 image--sentence pairs of bounding boxes and corresponding phrases, collected across eight different cardiopulmonary radiological findings, with an approximately equal number of pairs for each finding (see \cref{table:benchmark_stats_demographics}). It is curated to ensure gold-standard evaluation of phrase grounding.
The phrases in \dataset\ are not simple short captions, but genuine descriptions of radiological findings from original radiology reports \cite{johnson2019mimic} and dictated transcripts \cite{lanfredi2021reflacx}. Thus, compared to existing evaluation datasets, the proposed benchmark is a more challenging real-world image-text reasoning task. 

All the benchmark samples are chosen from the public MIMIC-CXR dataset \cite{goldberger2000physiobank,johnson2019mimic}. 
To collect a set of bounding-box labels, we first select samples from a set of studies with pre-existing image annotations (e.g., ellipses) \cite{lanfredi2021reflacx,tamliterati2020} and verify their correctness.
To link each image region with candidate phrases, we sampled sentences from the report of each study by extracting the highest matching sentences to the annotated labels using scores of the CheXbert classifier~\cite{smit2020combining}, and also used transcriptions of dictations when available~\cite{lanfredi2021reflacx}. 
Next, to better balance findings, we sampled additional studies at random as well as the ones used in the ImaGenome dataset \cite{wu2021chest}, the latter being a dataset of annotations of anatomical regions. Note that these sampled studies do not have preexisting region proposals. Radiologists then manually reviewed separate sets of candidates. If a bounding box was not available, the radiologists manually annotated the corresponding region(s) in the image with new bounding boxes. Radiologists rejected studies where no correct phrase candidates were available and where existing bounding boxes were placed incorrectly (e.g., covering too large an area). To ensure a high quality, consistent benchmark, the phrase-image samples that do not adhere to our guidelines (see Supp. \ref{appendix:ourdataset_label_collection}) were filtered out, such as phrases containing multiple abnormalities in distinct lung regions.

\vspace{-5pt}
\section{Experiments}\label{sec:experiments}
We conduct a comprehensive evaluation of our \cxrmodel\ language model as well as the proposed \jointmodel\ self-supervised VLP approach, and compare both to state-of-the art counterparts. \Cref{table:evaluation_types} shows how our evaluation coverage compares to recent related studies.
We begin by demonstrating \cxrmodel's superior performance and improved vocabulary, including on a radiology-specific NLI benchmark.
Next, we assess joint image-and-text understanding of \jointmodel\ on our new \dataset\ benchmark, which evaluates grounding of phrases describing radiological findings to the corresponding image regions.
We also investigate zero-shot classification and fine-tuning performance of \jointmodel\ on image- and pixel-level prediction tasks via the RSNA pneumonia dataset \cite{shih2019augmenting,wang2017chestx}.   

\begin{table}[tb]
\vspace{-6pt}
\centering
\caption{Comparing evaluations conducted in recent CXR image-text alignment studies.}
\setlength{\tabcolsep}{3pt}
\resizebox{.99\textwidth}{!}{
\begin{tabular}{@{}llcccccl@{}}
    \toprule
    \begin{tabular}{@{}l@{}}Downstream task\end{tabular} &
    \begin{tabular}{@{}l@{}}Used in ref.*\end{tabular} &
    \begin{tabular}{@{}l@{}}Image\\encoder\end{tabular} &
    \begin{tabular}{@{}l@{}}Text\\encoder\end{tabular} &
    \begin{tabular}{@{}l@{}}Phrase\\reasoning\end{tabular}&
    \begin{tabular}{@{}l@{}}Findings\\localisation \end{tabular}&
    \begin{tabular}{@{}l@{}}Latent\\alignment\end{tabular}&
    \begin{tabular}{@{}l@{}}Annotation\\availability\end{tabular} \\
\midrule
Natural language inference &[B] &  - & \checkmark & \checkmark & - & - & Scarce \\
Phrase grounding &[B] &  \checkmark & \checkmark & \checkmark &  \checkmark & \checkmark & Scarce \\
Image classification &[B,C,G,L,M] &  \checkmark & - & - & - & - & High \\
Zero-shot image classif. &[B,G] & \checkmark  & \checkmark &  -  & - & \checkmark & Moderate \\
\begin{tabular}{@{}l@{}}Dense image prediction \\ (e.g. segmentation)\end{tabular} &[B,G,L] &  \checkmark & - & - & \checkmark  & - & High \\
Global image--text retrieval &[C,G] &  \checkmark & \checkmark & - &  - & \checkmark & High \\
\bottomrule
\multicolumn{8}{@{}l@{}}{*B, \jointmodel\ (Proposed); C, ConVIRT~\cite{zhang2020contrastive}; G, GLoRIA~\cite{huang2021gloria}; L, LoVT~\cite{muller2021joint}; M, Local MI~\cite{liao2021multimodal}.}
\end{tabular}}
\label{table:evaluation_types}
\vspace{-10pt}
\end{table}
\vspace{-5pt}
\subsection{Setup}\label{sec:experiments_setup}
\paragraph{Datasets.}
We conduct experiments on the MIMIC-CXR v2~\cite{johnson2019mimic,goldberger2000physiobank} chest radiograph dataset, which provides 227,835 imaging studies with associated radiology reports for 65,379 patients, all collected in routine clinical practice. We only use frontal view scans (AP and PA) and also discard studies without an \Impression\ section. From this data, we establish a training set of 146.7k samples and a set of 22.2k validation samples, ensuring that all samples used for the different downstream evaluations are kept in a held-out test set. We emphasise that no labels are used during pretraining; for early stopping only a loss on validation data is tracked.
For evaluation, we use  RadNLI~\cite{miura2021improving} to assess the proposed \cxrmodel\ text model in isolation, the new \dataset\ assesses joint image--text understanding via phrase grounding, and the RSNA Pneumonia dataset~\cite{shih2019augmenting,wang2017chestx} to test zero-shot segmentation, as well as zero-shot and fine-tuned classification performance.

\vspace{-5pt}
\paragraph{Image and Text Pre-processing.} We downsize and centre crop images to a resolution of 512$\times$512 whilst preserving image aspect ratios. We perform image augmentations during training including: random affine transformations, random colour jitter, and horizontal flips (only for image fine-tuning tasks). 
For text model pre-training we utilise the `\Findings' and `\Impression' sections of reports, while joint training is performed using only the latter. 
During training, we perform sentence shuffling within sections as text-augmentation. Additionally, we perform limited automatic typo correction as in~\cite{chauhan2020joint}.

\vspace{-5pt}
\paragraph{Comparison Approaches.}
The proposed \cxrmodel\ text model is compared to the other specialised PubMedBERT~\cite{gu2021domain} and ClinicalBERT~\cite{alsentzer2019publicly} models. Note that ClinicalBERT was used in most related studies~\cite{huang2021gloria,liao2021multimodal,zhang2020contrastive,muller2021joint}. 
We compare \jointmodel\ to the closely related, state-of-the-art ConVIRT~\cite{zhang2020contrastive}, LoVT~\cite{muller2021joint} and GLoRIA~\cite{huang2021gloria} approaches (see \cref{sec:relatedwork}). Lastly, we create \localjointmodel\ by extending \jointmodel\ with the local loss term introduced in \cite{huang2021gloria} to illustrate the complementary role of proposed pre-training strategy to recent advances in biomedical VLP.

\vspace{-5pt}
\paragraph{Metrics.}
We report segmentation results via mean intersection over union (mIoU) and contrast-to-noise ratio (CNR), and report the Dice score~\cite{crum2006generalized} to compare to~\cite{muller2021joint}. We first compute the cosine similarity between a projected phrase embedding $\cls$ and local image representations $\localrep$, resulting in a grid of scores between $[-1,1]$. The similarities are later thresholded to compute mIoU and Dice score. The mIoU is defined as an average over the thresholds [0.1, 0.2, 0.3, 0.4, 0.5].
The CNR measures the discrepancy between scores inside and out of the bounding box region, without requiring hard thresholds.
This evaluation of local similarities is important as some clinical downstream applications may benefit from heatmap visualisations as opposed to discrete segmentations.
For CNR, let $A$ and $\overline{A}$ denote the interior and exterior of the bounding box, respectively. We then compute $\operatorname{CNR}=|\mu_A-\mu_{\overline{A}}|/(\sigma_A^2+\sigma_{\overline{A}}^2)^{\frac12}$, where $\mu_X$ and $\sigma_X^2$ are the mean and variance of the similarity values in region $X$. 
\vspace{-2pt}
\subsection{Text Model Evaluation}
\paragraph{Natural Language Understanding.}
We use the RadNLI benchmark~\cite{miura2021improving} to evaluate how well the proposed \cxrmodel\ text model captures domain-specific semantics. The dataset contains labelled hypothesis and premise pairs, sourced from MIMIC-CXR radiology reports, with the following label categories: (1)~entailment, i.e.\ the hypothesis can be inferred from the premise; (2)~contradiction, i.e.\ the hypothesis cannot be inferred from the premise; and (3)~neutral, i.e.~the inference relation is undetermined.
RadNLI provides expert-annotated development and test sets (480 examples each), but no official training set. Thus, following \cite{miura2021improving}, we use MedNLI~\cite{shivade2019mednli} for training, which has 11k samples sourced from MIMIC-III discharge summaries, with equally distributed NLI labels.
We fine-tune the language models up to 20 epochs and use early stopping by monitoring accuracy scores on the RadNLI development set.
\Cref{table:text_encoder} summarises the NLI evaluation, masked token prediction, and subword tokenisation results. Using only MedNLI training samples, our model achieves a good accuracy of $65.21\%$, and far outperforms fine-tuned ClinicalBERT, PubMedBERT, and the score reported in RadNLI~\cite{miura2021improving}. Another important result is that RadNLI accuracy improves after joint training with images (last row of \cref{table:text_encoder}).

\begin{table}[tb]
\vspace{-6pt}
\centering
\caption{Evaluation of text encoder intrinsic properties and fine-tuning for radiology natural language inference: (1)~RadNLI fine-tuning scores (average of 5 runs); (2)~Mask prediction accuracy on MIMIC-CXR val.\ set; (3)~Vocabulary comparison, number of tokens vs.\ original number of words in \Findings, increase shown as percentage.}
\setlength{\tabcolsep}{5pt}
\resizebox{.9\textwidth}{!}{
\begin{tabular}{@{}lcccc@{}}
    \toprule
  & \multirow{2}{*}{\shortstack[c]{RadNLI accuracy \\ (MedNLI transfer)}} &  \multirow{2}{*}{\shortstack[c]{Mask prediction \\\strut accuracy}}  & \multirow{2}{*}{\shortstack[c]{Avg. \texttt{\#} of tokens\\after tokenization}} &  \multirow{2}{*}{\shortstack[c]{Vocabulary \\ size}} \\
  \\
\midrule
RadNLI baseline~\cite{miura2021improving}  & 53.30 &  -  & -  & - \\
ClinicalBERT  &  47.67 & 39.84  & 78.98 (\Plus38.15\%) & 28,996 \\
PubMedBERT  &  57.71  & 35.24   & 63.55 (\Plus11.16\%)  & 28,895 \\
\midrule
\cxrmodel\ (after Phase-III) &  60.46 & 77.72   & 58.07 (\Plus1.59\%) & 30,522 \\
\cxrmodel\ (after Phase-III + Joint Training)  &  65.21 & 81.58   & 58.07 (\Plus1.59\%) & 30,522 \\
\bottomrule
\end{tabular}}{}
\label{table:text_encoder}
\vspace{-15pt}
\end{table}

\vspace{-6pt}
\paragraph{Mask Prediction Accuracy.} While mask prediction accuracy does not always translate to downstream application performance, it is an auxiliary metric that captures important aspects of a language model's grasp of a target domain. We report Top-1 mask prediction accuracy on radiology reports in the MIMIC-CXR validation set (\cref{table:text_encoder}), and follow the standard masking configuration (15\% masking probability). Despite being trained on closely related data, our \cxrmodel\ displays a much better mask prediction accuracy compared to ClinicalBERT (trained on MIMIC-III, which includes radiology reports) and PubMedBERT (trained on biomedical literature text). This suggests that radiology text significantly differs from other clinical text or biomedical literature text, highlighting the need for specialised text encoder models.

\begin{wraptable}{r}{6.5cm}
\vspace{-23pt}
\centering
\caption{\cxrmodel\ ablation.
CNR and mIoU are macro averages of \jointmodel\ performance on all categories of \dataset.
\textit{Syn.~sim.} denotes the average cosine similarity between RadNLI entailments.
\textit{Cont.~gap} is the average similarity gap of RadNLI entailment and contradiction pairs. 
\cxrmodel\ is the combination of all components below the first row.
}
\setlength{\tabcolsep}{5pt}
\resizebox{\linewidth}{!}{
\begin{tabular}{@{}lccccccc@{}}
\toprule
  & \multicolumn{2}{c}{RadNLI} & \multicolumn{2}{c}{Grounding} \\
  \cmidrule(lr){2-3} \cmidrule(lr){4-5}
  Model or pretraining stage & Syn.~sim. & Cont.~gap & mIoU & CNR \\
\midrule
 ClinicalBERT                       & .657 & .609 & .182 & 0.791 \\ %
 \midrule
 Pretrain \& Vocab  (I--II)         & .749 & .646 & .194 & 0.796 \\ %
 + MLM loss added to joint training       & .871 & .745 & .209 & 0.860 \\ %
 + Use of attention drop-out (III)  & .893 & .802 & .217 & 0.945 \\ %
 + RSM Pretrain (III)      & .877 & .779 & .220 & 1.012 \\ %
 + Sentence shuffling (\cxrmodel)   & .884 & .798 & .220 & 1.031 \\ %
\bottomrule
\end{tabular}}{}
\label{table:ablation_study}
\vspace{-15pt}
\end{wraptable}

\vspace{-6pt}
\paragraph{Ablation.}
We also conduct an ablation of the various aspects of \cxrmodel, measuring the impact after joint training. \Cref{table:ablation_study} shows that all components of \cxrmodel\ contribute to improved downstream and NLI performance, both in terms of alignment between related sentences (entailments) and of discrimination of contradictions. In particular, note the substantial improvement on these scores due to keeping the MLM objective during joint finetuning.

\subsection{Local Alignment Evaluation -- Phrase Grounding}\label{sec:phrase_grounding}
We perform a phrase grounding evaluation of the pretrained \jointmodel\ model on the \dataset\ dataset. For each image--phrase pair, the image is passed to the CNN image encoder and projected to obtain a grid of image representations $\localrep$ in the joint space. Similarly, the phrase is embedded via the text encoder and projected to the joint space to obtain $\cls$. Cosine similarity between $\cls$ and elements of $\localrep$ produces a similarity grid, which is evaluated against the ground-truth bounding boxes.
\Cref{table:phrasegrounding} shows the superior phrase grounding results achieved by \jointmodel\ across radiological findings and further shows that the addition of local losses as in our \localjointmodel\ can improve phrase grounding performance for almost all findings.
Moreover, the ablation in \cref{table:ablation_study} demonstrates that there are clear gains to be had in visual grounding performance by improving the text model.

\begin{table}[tb]
\vspace{-6pt}
\caption{Contrast-to-noise ratio (CNR) obtained on the newly released \dataset\ dataset, averaged over four runs with different seeds. The results are collected using different text encoder and training objectives (e.g., G\&L: Global and local loss).}
\resizebox{.99\textwidth}{!}{
\begin{tabular}{@{}lllccccccccc@{}}
    \toprule
 Method & Objective & Text encoder                          & Atelectasis & Cardiomegaly & Consolidation & Lung opacity & Edema & Pneumonia & Pneumothorax & Pl. effusion & Avg.\\
\midrule
Baseline                            & Global & ClinicalBERT &  0.70$\pm$.03 & 0.53$\pm$.04 & 1.15$\pm$.07 & 0.75$\pm$.12 & 0.83$\pm$.04 & 0.85$\pm$.09 & 0.29$\pm$.01 & 1.05$\pm$.05 & 0.769$\pm$.02\\
Baseline                            & Global & PubMedBERT   &  0.72$\pm$.08 & 0.64$\pm$.05 & 1.22$\pm$.07 & 0.69$\pm$.07 & 0.80$\pm$.04 & 0.91$\pm$.09 & 0.21$\pm$.07 & 0.99$\pm$.03 & 0.773$\pm$.05\\
ConVIRT \cite{zhang2020contrastive} & Global & ClinicalBERT &  0.86$\pm$.04 & 0.64$\pm$.06 & 1.25$\pm$.06 & 0.78$\pm$.07 & 0.68$\pm$.07 & 1.03$\pm$.05 & 0.28$\pm$.08 & 1.02$\pm$.03 & 0.818$\pm$.01\\
GLoRIA \cite{huang2021gloria}       & G\&L & ClinicalBERT   &  0.98$\pm$.04 & 0.53$\pm$.31 & 1.38$\pm$.03 & 1.05$\pm$.04 & 0.66$\pm$.03 & 1.18$\pm$.04 & 0.47$\pm$.02 & 1.20$\pm$.04 & 0.930$\pm$.03\\
\midrule
\jointmodel                         & Global & \cxrmodel    &  1.02$\pm$.06 & 0.63$\pm$.08 & 1.42$\pm$.02 & 1.05$\pm$.06 & 0.93$\pm$.03 & 1.27$\pm$.04 & 0.48$\pm$.06 & 1.40$\pm$.06 & 1.027$\pm$.02\\
\localjointmodel                         & G\&L & \cxrmodel &  1.17$\pm$.04 & 0.95$\pm$.21 & 1.45$\pm$.03 & 1.19$\pm$.05 & 0.96$\pm$.05 & 1.19$\pm$.01 & 0.74$\pm$.05 & 1.50$\pm$.03 & 1.142$\pm$.04\\
\bottomrule
\end{tabular}}{}
\label{table:phrasegrounding}
\end{table}

\begin{table}[t]
\vspace{-10pt}
\centering
\caption{RSNA Pneumonia zero-shot and fine-tuned classification. We compare to GLoRIA scores reported in \cite{huang2021gloria}  which outperforms ConVIRT~\cite{zhang2020contrastive} (see ~\cite{huang2021gloria}). Training size: GLoRIA ($N=186k$, private dataset), \jointmodel\ ($N=146.7k$ of MIMIC-CXR).}\label{table:rsna_classification}
\begin{minipage}[c]{0.38\linewidth}
\centering
\includegraphics[width=\linewidth,clip,trim={6pt 6pt 6pt 6pt}]{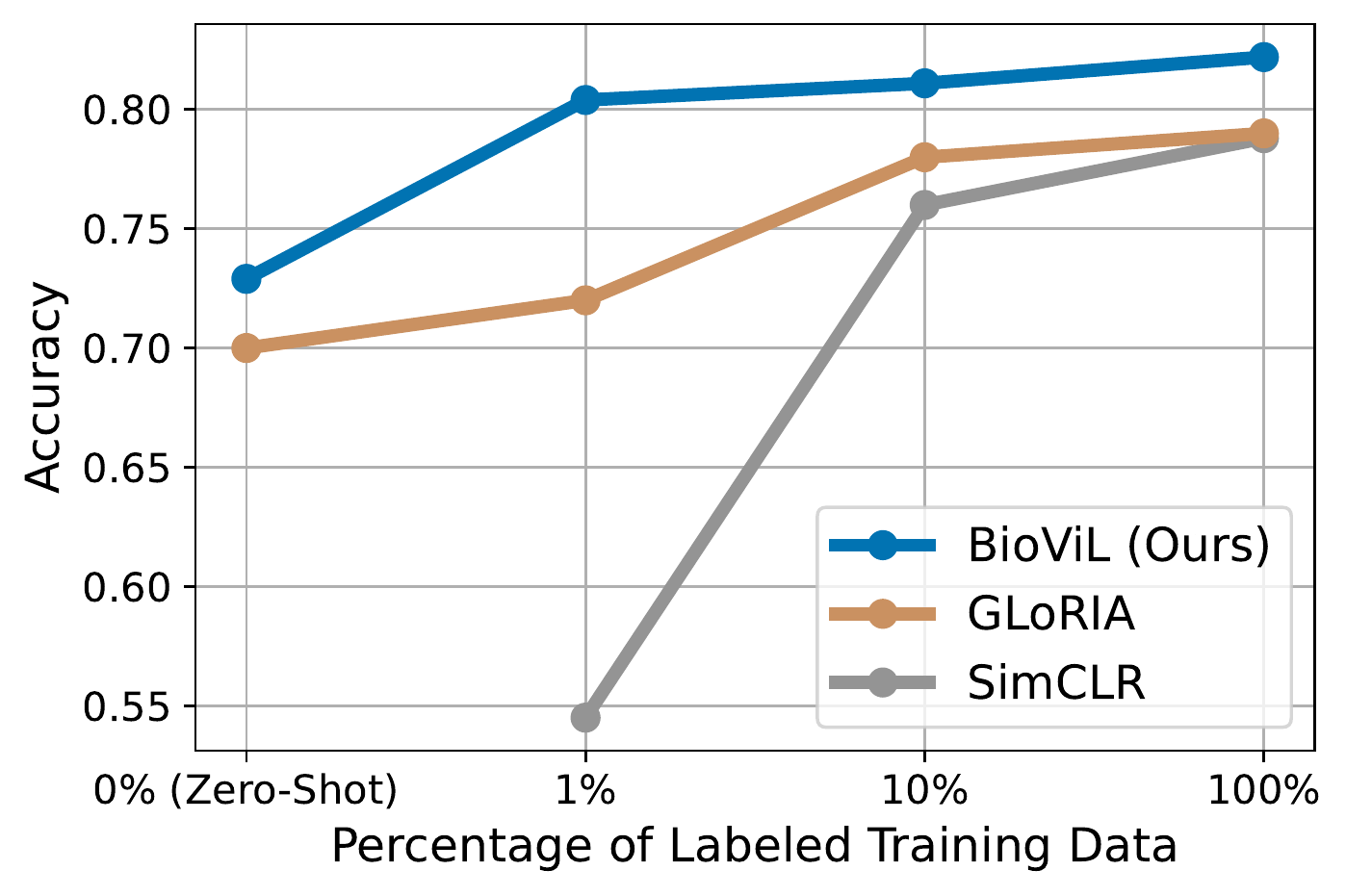}
\end{minipage}\hfill%
\begin{minipage}[c]{0.6\linewidth}
\centering
\scriptsize
\setlength{\tabcolsep}{3pt}
\resizebox{\linewidth}{!}{
\begin{tabular}{@{}llclcccc@{}}
    \toprule
Method & Type & Text model & Loss & \% of labels & Acc. & F1 & AUROC \\
\midrule
\multirow{3}{6em}{SimCLR~\cite{sicmlr}} & \multirow{3}{3.5em}{Image only} & \multirow{3}{*}{-} & \multirow{3}{*}{Global} & 1\% & 0.545 & 0.522 & 0.701 \\
 &&&& 10\% & 0.760 & 0.639 & 0.802 \\
 &&&& 100\% & 0.788 & 0.675 & 0.849 \\
 \midrule
\multirow{4}{6em}{GLoRIA~\cite{huang2021gloria}} & \multirow{4}{*}{Joint} & \multirow{4}{*}{ClinicalBERT}& \multirow{4}{3.5em}{Global \& local}
 & Zero-shot & 0.70 & 0.58 & - \\
 &&&& 1\% & 0.72 & 0.63 & 0.861 \\
 &&&& 10\% & 0.78 & 0.63 & 0.880 \\
 &&&& 100\% & 0.79 & 0.65 & 0.886 \\
 \midrule
 Baseline & Joint & ClinicalBERT & Global & Zero-shot & 0.719 & 0.614 & 0.812  \\
 \midrule
\multirow{4}{*}{\jointmodel} & \multirow{4}{*}{Joint} & \multirow{4}{*}{CXR-BERT}  & \multirow{4}{*}{Global} &  Zero-shot & 0.732 & 0.665 & 0.831 \\
 &&  &  & 1\% & 0.805 & 0.723 & 0.881 \\ 
 &&  &  & 10\% & 0.812 & 0.727 & 0.884 \\ 
 &&  &  & 100\% & 0.822& 0.733 & 0.891 \\ 
\bottomrule
\end{tabular}}
\end{minipage}
\vspace{-15pt}
\end{table}

\subsection{Global Alignment Evaluation -- Zero-shot \& Linear Probing}\label{sec:rsnaclassification}
\vspace{-4pt}
To measure global alignment quality, the joint models are also benchmarked on zero-/few-shot binary pneumonia classification problems (image-level) using the external RSNA dataset~\cite{shih2019augmenting}. Fine-tuning is done via linear probing, i.e. only a last linear layer is trained. 
The evaluation is conducted on $\mathcal{D}_\mathrm{test}=9006$ images as in \cite{huang2021gloria} (30\% eval.\ / 70\% train.) using the dataset's ground-truth labels. We define two simple text prompts for 
\jointmodel, representing presence/absence of pneumonia: ``Findings suggesting pneumonia'' and ``No evidence of pneumonia''. 
The image encoders are utilised and fine-tuned as described in \cref{sec:jointmodel}. 

The zero-shot and fine-tuned results in \Cref{table:rsna_classification} show that our focus on better text modelling results in improved joint modelling of shared latent information between text-image pairs. Note that, to achieve its superior performance here and in \cref{sec:rsnalocalisation}, \jointmodel\ does not require extensive human expert text-prompt engineering (see Supp.~\ref{sec:textpromptsensitivity} for a sensitivity analysis) as for example conducted in GLoRIA~\cite{huang2021gloria}, where variations over severity and/or location were created.

\begin{wraptable}{r}{6.5cm}
\vspace{-24pt}
\centering
\small
\caption{RSNA pneumonia segmentation, showing \textit{Zero-shot} and \textit{linear probing} results. Related work is reproduced in the same experimental setup except for LoVT \cite{muller2021joint}.}
\scriptsize
\resizebox{\linewidth}{!}{
\begin{tabular}{@{}lclccc@{}}
    \toprule
    Method & \% of Labels & Supervision & IoU & Dice & CNR\\
    \midrule
    LoVT \cite{muller2021joint}         & 100\%  & Lin. prob. & - & 0.518 & - \\
    ConVIRT \cite{zhang2020contrastive} & -            & Zero-shot   & 0.228 & 0.348 & 0.849 \\
    GLoRIA \cite{huang2021gloria}       & -            & Zero-shot   & 0.245 & 0.366 & 1.052 \\
    \jointmodel                         & -            & Zero-shot   & 0.355 & 0.496 & 1.477 \\
    \midrule
    SimCLR \cite{sicmlr}      & 5\%    & Lin. prob. & 0.382 & 0.525 & 1.722 \\
    SimCLR \cite{sicmlr}      & 100\%  & Lin. prob. & 0.427 & 0.570 & 1.922 \\
    \jointmodel                       & 5\%    & Lin. prob. & 0.446 & 0.592 & 2.077 \\
    \jointmodel                       & 100\%   & Lin. prob. & 0.469 & 0.614 & 2.178 \\
    \bottomrule
\end{tabular}
}{}
\label{table:zero_shot_localization}
\vspace{-20pt}
\end{wraptable}

\subsection{Local Alignment Evaluation -- Semantic Segmentation}\label{sec:rsnalocalisation}
We evaluate models on an RSNA pneumonia segmentation task, using grid-level image representations in the joint latent space.
We use the same text prompts as in the previous section for all models, and evaluate against ground-truth bounding boxes of the RSNA pneumonia dataset ($|\mathcal{D}_\mathrm{train}|=6634$ and $|\mathcal{D}_\mathrm{test}|=2907$).  \Cref{table:zero_shot_localization} shows that \jointmodel\ significantly reduces the need for dense annotations as compared to similar multi-modal and image-only pretraining approaches, outperforming them when using the same number of labelled data points. 
Note that our proposed modelling framework \jointmodel (\cref{fig:mainssvlp}),
uses neither a local loss term \cite{huang2021gloria,muller2021joint}, nor a separate object detection \cite{redmon2018yolov3} or segmentation network \cite{ronneberger2015u}.
Further, while \Cref{table:zero_shot_localization} shows results using two simple queries, we find that \jointmodel\ continues to outperform related work even when more prompts are used for all models as in \cite{huang2021gloria}.
Dice and IoU are computed using the same threshold of $0.6$ on predictions scaled between $[0,1]$.

\vspace{-5pt}
\section{Related Work}\label{sec:relatedwork}
\vspace{-5pt}
We refer the reader to Supp.~\ref{sec:app_relatedwork} for a more detailed review of related work. \\[2 mm]
\textit{Biomedical Vision--Language Processing.} Multiple studies explore joint representation learning for paired image and text radiology data~\cite{hsu2018unsupervised,huang2021gloria,liao2021multimodal,muller2021joint,zhang2020contrastive}. \cite{zhang2020contrastive} follow a contrastive learning formulation for instance-level representation learning, while~\cite{huang2021gloria,muller2021joint} introduce approaches that combine instance-level radiology image--report learning with local terms.
An alternative, local-only objective is explored by~\cite{liao2021multimodal}, approximating the mutual information between local image features and sentence-level text features.
While most related approaches use no ground truth, \cite{chauhan2020joint} study a semi-supervised edema severity classification setting,  and \cite{hayat2021multi} assume sets of seen and unseen labels towards CXR zero-shot classification.

Related medical VLP work commonly uses publicly available contextual word embedding models including BioBERT~\cite{lee2020biobert}, ClinicalBERT~\cite{alsentzer2019publicly},  BioClinicalBERT~\cite{alsentzer2019publicly}, or PubMedBERT~\cite{gu2021domain}. The models are either trained from scratch or fine-tuned via continual pretraining using an MLM objective. Additional objectives such as adversarial losses~\cite{liu2020adversarial} are added infrequently. The specialised corpora these 
models use include PubMed abstracts and PubMed Central full texts (see \cite{alsentzer2019publicly}), as well as MIMIC-III~\cite{johnson2016mimic} clinical notes. \\[2 mm]
\textit{Local Alignment Datasets.} Presently, no datasets exist that allow for phrase grounding of radiology findings, but some enable different forms of local image evaluations. VinDr~\cite{nguyen2020vindr}, RSNA Pneumonia~\cite{shih2019augmenting}, and the NIH Chest X-ray Dataset~\cite{wang2017chestx} provide bounding-box annotations, but lack free-text descriptions. REFLACX~\cite{lanfredi2021reflacx} provides gaze locations (ellipses) captured with an eye tracker, dictated reports, and some ground truth annotations for gaze locations, but no full phrase matches to image regions. Phrase annotations for MIMIC-CXR data released in~\cite{tamliterati2020} are of small size (350 studies), only contain two abnormalities, and for some samples have shortened phrases that were adapted to simplify the task. %
The ground-truth set of ImaGenome~\cite{wu2021chest} only contains 500 studies, bounding-box regions annotate anatomical regions rather than radiological findings, and its sentence annotations are not curated for grounding evaluation.

\section{Conclusion}
We show that careful attention to text modelling can lead to large benefits for all learned models in self-supervised vision language processing (VLP) frameworks for medical applications. 
We introduce a novel pretraining procedure and publicly release a radiology domain-specific language model: \cxrmodel. It has an improved vocabulary and understanding of radiology sentences, contributing to improved downstream performance for all aspects of VLP approaches, e.g., the superior performance on a radiology natural language inference benchmark.

We also present \jointmodel, as a simple yet effective baseline for self-supervised multi-modal learning for paired image--text radiology data, with a focus on improved text modelling. 
The approach displays state-of-the-art performance on a large number of downstream tasks evaluating global and local aspects of the image model, text model, and joint latent space. On zero-shot tasks, the model does not require extensive text-prompt engineering compared to prior work. Notably, it outperforms related work on segmentation without requiring a local loss term or an additional vision model to produce region proposals. In that regard, it is complementary to local contrastive losses, and the combination of the two yields improved phrase grounding performance (\cref{table:phrasegrounding}).

To support the research community in evaluating fine-grained image--text understanding in the radiology domain, we also publicly release a chest X-ray phrase grounding dataset called \dataset. It presents a more challenging benchmark for joint image--text understanding compared to existing datasets, requiring reasoning over real-world radiology language and scans to ground findings in the correct image locations. Limitations of the proposed joint approach include that it does not explicitly deal with false negatives in the contrastive losses. 
Furthermore, co-occurrence of multiple abnormalities could enable contrastive methods to focus only on a subset to match pairs, e.g. pneumothorax and chest tubes commonly occur together~\cite{graf2020pneumothorax}. 
Amongst its failure cases (see Supp. \ref{appendix:failurecases} for more), we have seen that the approach struggles with very small structures, likely due to image resolution limits. 
Future work will expand the evaluated radiological findings, and explore using larger image resolution. \\[2mm]
\textbf{Acknowledgements:} We would like to thank Dr Javier Gonz\'alez and Fernando P\'erez-Garc\'ia for their valuable feedback and contributions, Hannah Richardson for helping  with the compliance review of the datasets, and Dr Matthew Lungren for their clinical input and data annotations provided to this study.

\bibliographystyle{splncs04}
\bibliography{references}

\beginsupplement
\appendix

\section{Additional Experiments}

\subsection{Zero-shot Text-prompt Sensitivity Analysis}\label{sec:textpromptsensitivity}
Vision-language pretraining aligns image and text data in a joint representation space, which enables impressive zero-shot downstream image classification performance via input text prompts. However, some recent work~\cite{huang2021gloria,zhang2020contrastive} has shown that downstream task performance can heavily depend on the choice of text prompts. Constructing good text prompts (prompt engineering) may require expert domain knowledge and can be costly and time-consuming. In \cref{tab:text_prompt_sensitivity}, we study RSNA pneumonia zero-shot classification performance using different text prompt combinations. Compared to the baseline, \jointmodel ~demonstrates much lower sensitivity to prompt choices selected from the data distribution. \jointmodel ~maintains its high performance even when faced with relatively long queries, which is not the case for the baseline model. These observations suggest that our improved text encoder \cxrmodel ~is more robust to prompt variations, and makes prompt engineering easier and less of a requirement to achieve high zero-shot classification performance.

\begin{table}
    \vspace{-15pt}
    \centering
    \caption{Text prompt sensitivity analysis on the RSNA pneumonia zero-shot classification task. Image-text models trained without the proposed text modelling improvements (\cref{table:ablation_study}) show higher sensitivity to different input text prompts as the latent text embeddings are  inconsistent for synonym phrases. For this reason, baseline methods often require post-hoc text prompt engineering heuristics (e.g. \cite{huang2021gloria}).}
    \label{tab:text_prompt_sensitivity}
    \scriptsize
    \setlength{\tabcolsep}{.5em}
    \newcommand{\rowskip}{\rule{0pt}{3ex}}
    \resizebox{.95\textwidth}{!}{
    \begin{tabular}{@{} l >{``}p{5.8cm}<{"} >{``}p{5.2cm}<{"} ccc @{}}
        \toprule
        Method & \multicolumn{1}{@{}l@{}}{Pos. Query} & \multicolumn{1}{@{}l@{}}{Neg. Query} & F1 Score & ROC-AUC  & $|\Delta AUC| $ \\
        \midrule
        \jointmodel & Findings suggesting pneumonia & There is no evidence of acute pneumonia   &  0.657    & 0.822 & -\\\rowskip
        ClinicalBert & Findings suggesting pneumonia & There is no evidence of acute pneumonia  & 0.581     & 0.731 & -\\
        \midrule
        \jointmodel & Findings suggesting pneumonia & No evidence of pneumonia                  & 0.665     & 0.831 & -               \\\rowskip
        \jointmodel & Consistent with the diagnosis of pneumonia & There is no evidence of acute pneumonia  & 0.669 &  0.839 & 0.008  \\\rowskip
        ClinicalBert & Findings suggesting pneumonia & No evidence of pneumonia                 & 0.614     & 0.815 & -               \\\rowskip
        ClinicalBert & Consistent with the diagnosis of pneumonia & There is no evidence of acute pneumonia &  0.621 &  0.694 & 0.121 \\ 
        \midrule
         \jointmodel & Findings consistent with pneumonia & No evidence of pneumonia            & 0.672     & 0.838 & -     \\\rowskip
        \jointmodel & Findings consistent with pneumonia & There is no pneumonia                & 0.679     & 0.847 & 0.009 \\\rowskip
        ClinicalBert & Findings consistent with pneumonia & No evidence of pneumonia            & 0.640     & 0.782 & -     \\\rowskip
        ClinicalBert & Findings consistent with pneumonia & There is no pneumonia               & 0.586     & 0.724 & 0.058 \\
        \bottomrule
    \end{tabular}}
    \vspace{-15pt}
\end{table}

\subsection{Qualitative Results -- Phrase Grounding}\label{appendix:failurecases}
\begin{figure}[p]
    \centering
    \begin{subfigure}[b]{0.9\linewidth}
        \includegraphics[width=\textwidth]{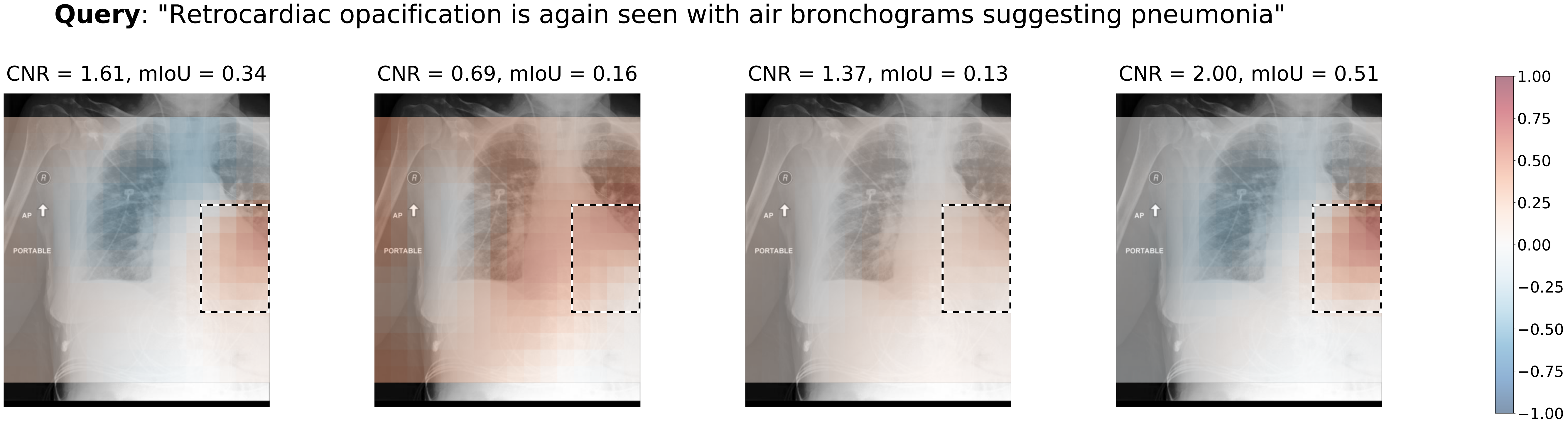}
        \caption{Relatively long and complex query}
    \end{subfigure} \\[1ex]
    \begin{subfigure}[b]{0.9\linewidth}
        \includegraphics[width=\textwidth]{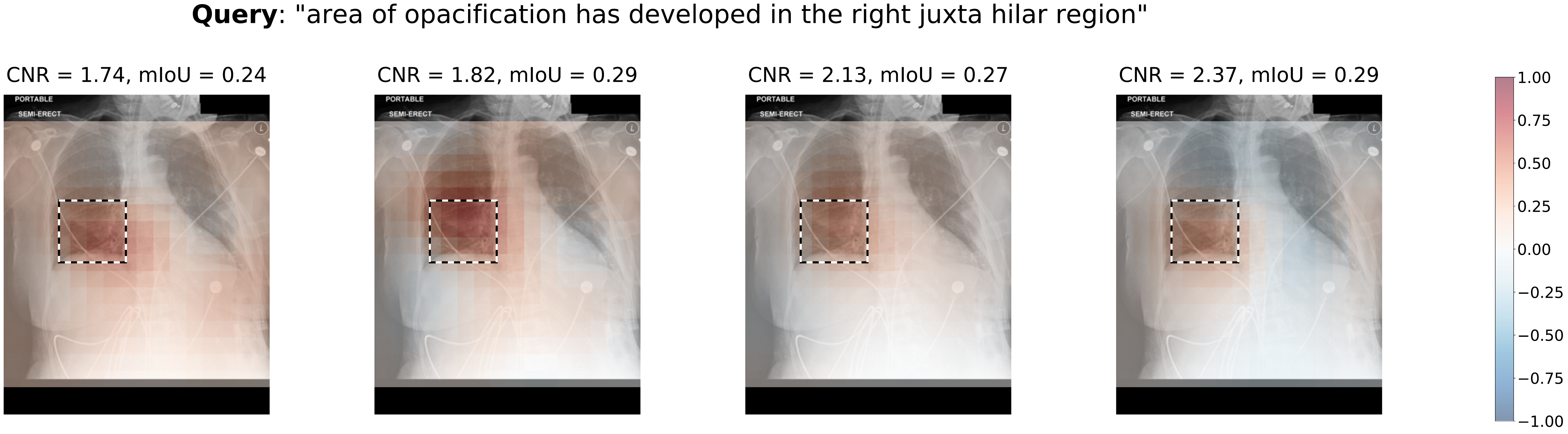}
        \caption{Complex anatomical location specification}
    \end{subfigure} \\[1ex]
    \begin{subfigure}[b]{0.9\linewidth}
        \includegraphics[width=\textwidth]{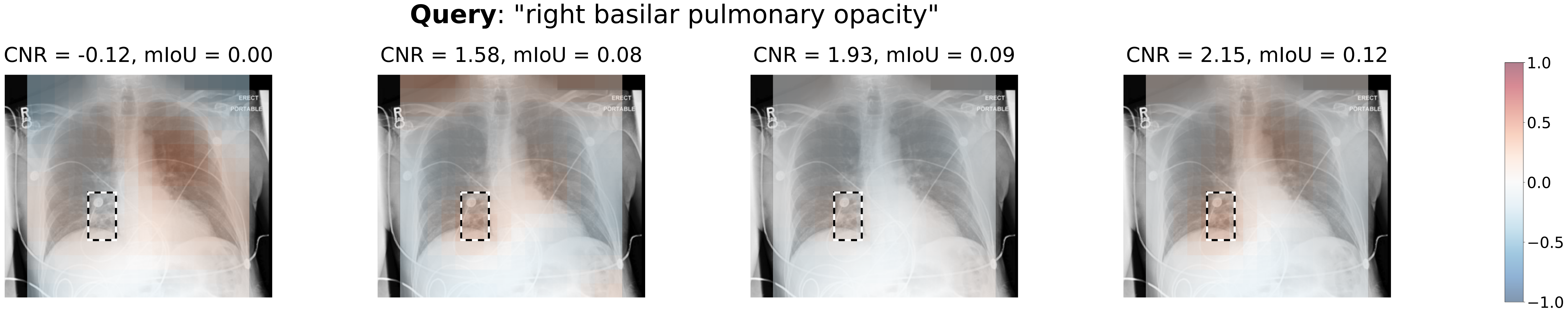}
        \caption{Small ground-truth bounding box}
    \end{subfigure} \\[1ex]
    \begin{subfigure}[b]{0.9\linewidth}
        \includegraphics[width=\textwidth]{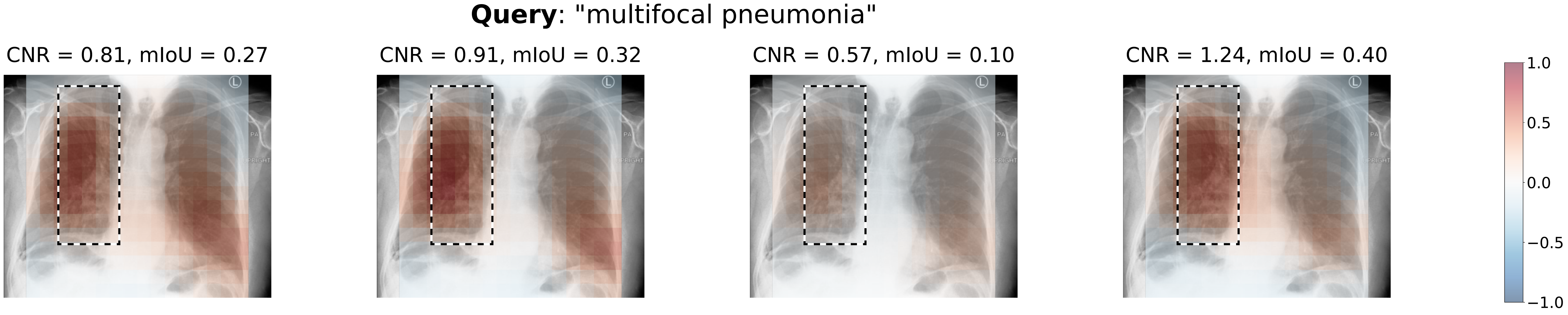}
        \caption{Multifocal pneumonia example which is localised in the right lobe}
    \end{subfigure} \\[1ex]
    \begin{subfigure}[b]{0.9\linewidth}
        \includegraphics[width=\textwidth]{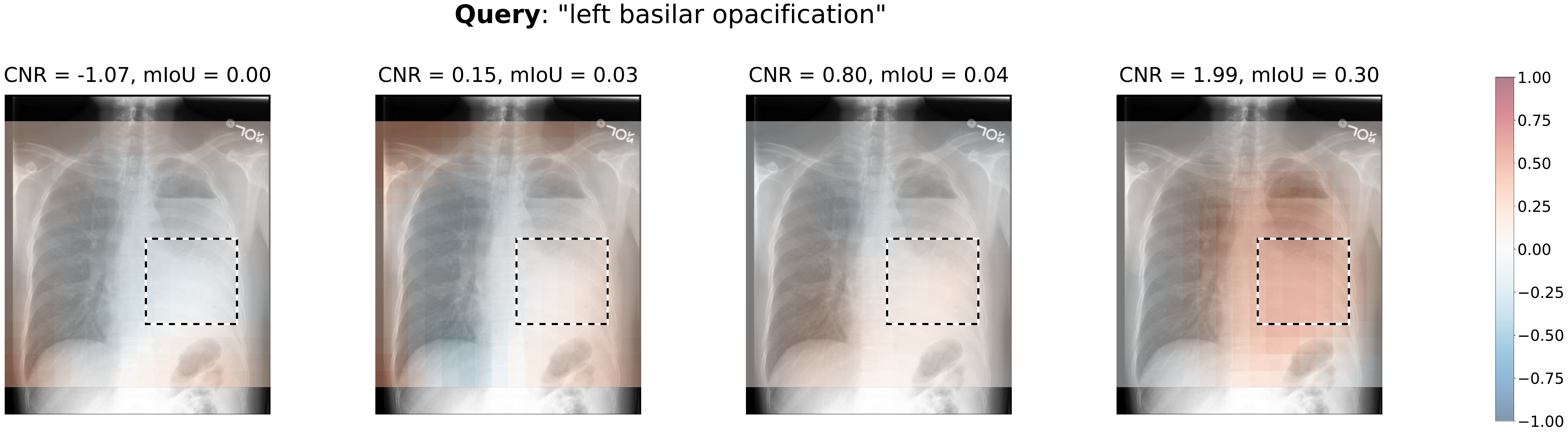}
        \caption{Location modifier ``left basilar''}
    \end{subfigure}
    \caption{Qualitative examples from \dataset~phrase grounding benchmark. Model outputs (latent  vector  similarity) are compared (from left, ClinicalBERT baseline, ConVIRT, GLoRIA, and \jointmodel)}
    \label{fig:qualitative_examples}
\end{figure}

In \cref{fig:qualitative_examples}, we show and describe some phrase grounding examples obtained with different models on the \dataset\ dataset. From left to right, the figure shows the ClinicalBERT baseline, ConVIRT, GLoRIA, and \jointmodel\ similarity maps. While the figure only illustrates a few examples, the results demonstrate that phrase grounding performance can be significantly enhanced by leveraging improved text modelling (\jointmodel). The examples include clinical findings that differ in size, type, and anatomical location.

\begin{figure}[p]
    \centering
    \begin{subfigure}[b]{0.9\linewidth}
        \includegraphics[width=\textwidth]{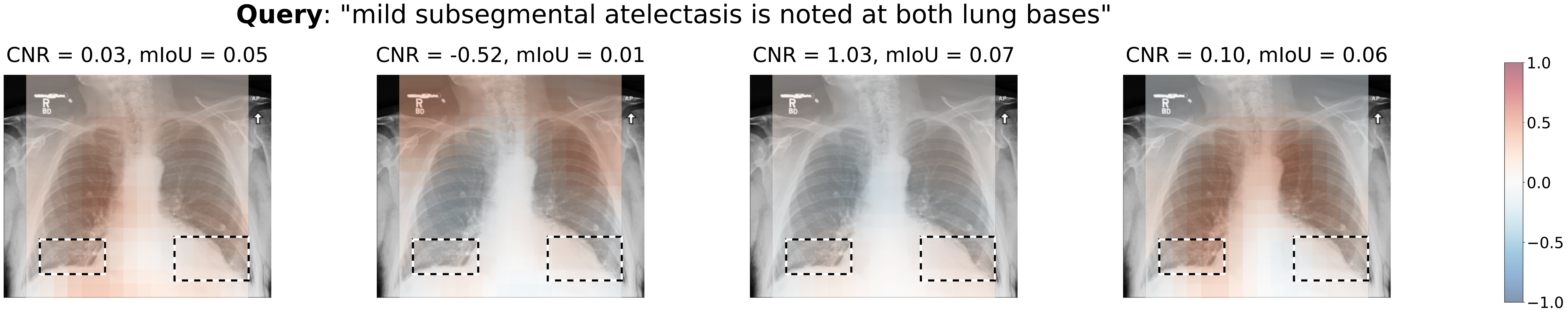}
        \caption{Failed to recognise atelectasis despite having ``both lung bases'' location specification}
    \end{subfigure} \\[1ex]
    \begin{subfigure}[b]{0.9\linewidth}
        \includegraphics[width=\textwidth]{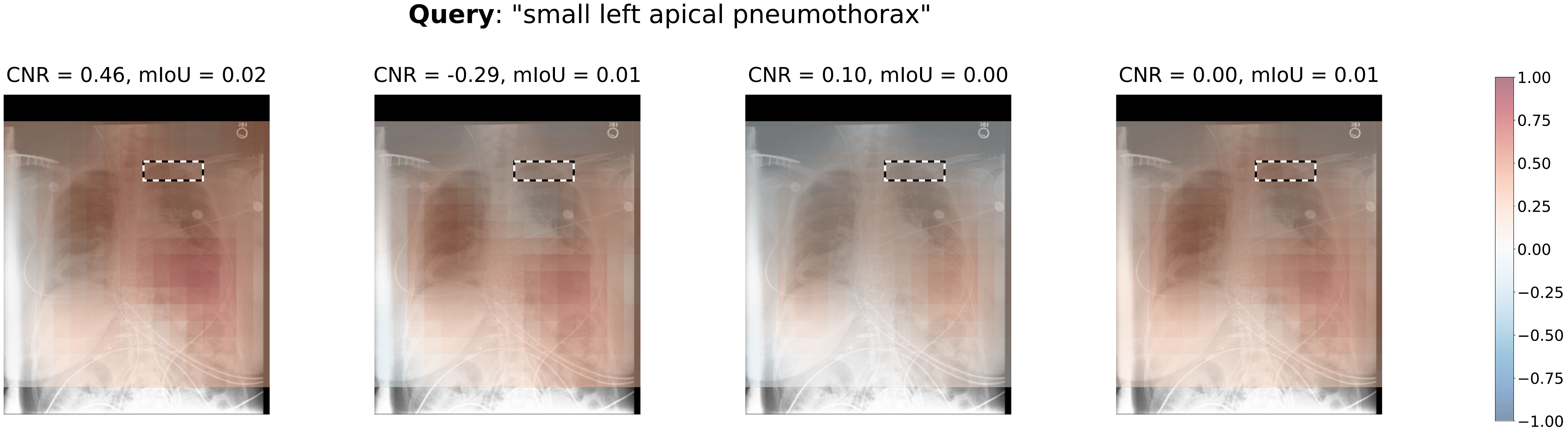}
        \caption{Failed to recognise small pneumothorax despite having ``apical'' location modifier}
    \end{subfigure} \\[1ex]
    \begin{subfigure}[b]{0.9\linewidth}
        \includegraphics[width=\textwidth]{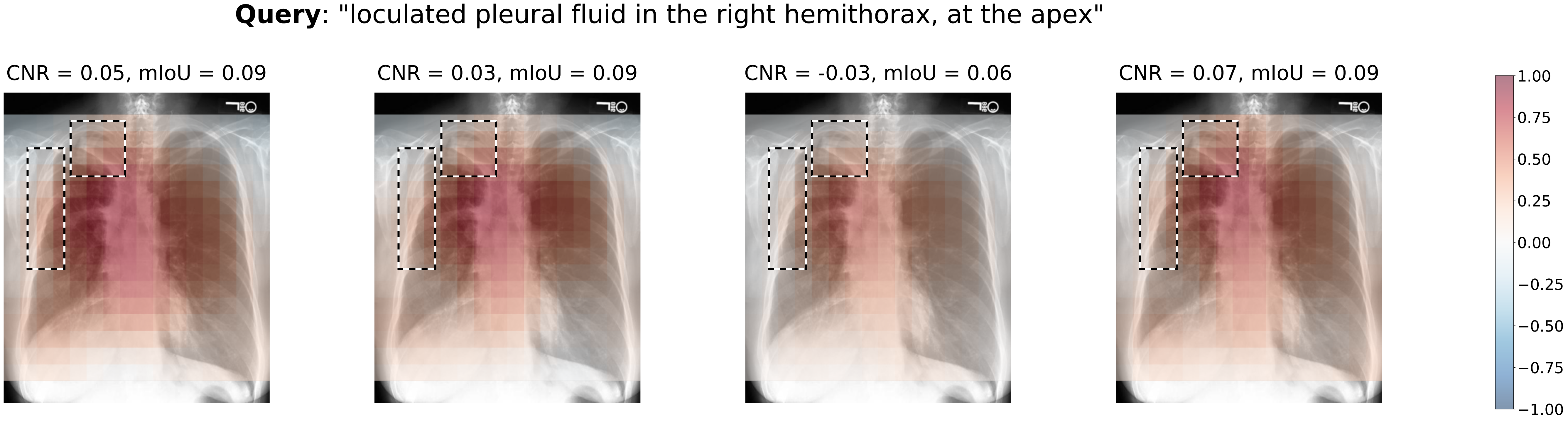}
        \caption{Failed to recognise loculated pleural fluid despite having ``apical'' and ``right hemithorax''}
    \end{subfigure} \\[1ex]
    \begin{subfigure}[b]{0.9\linewidth}
        \includegraphics[width=\textwidth]{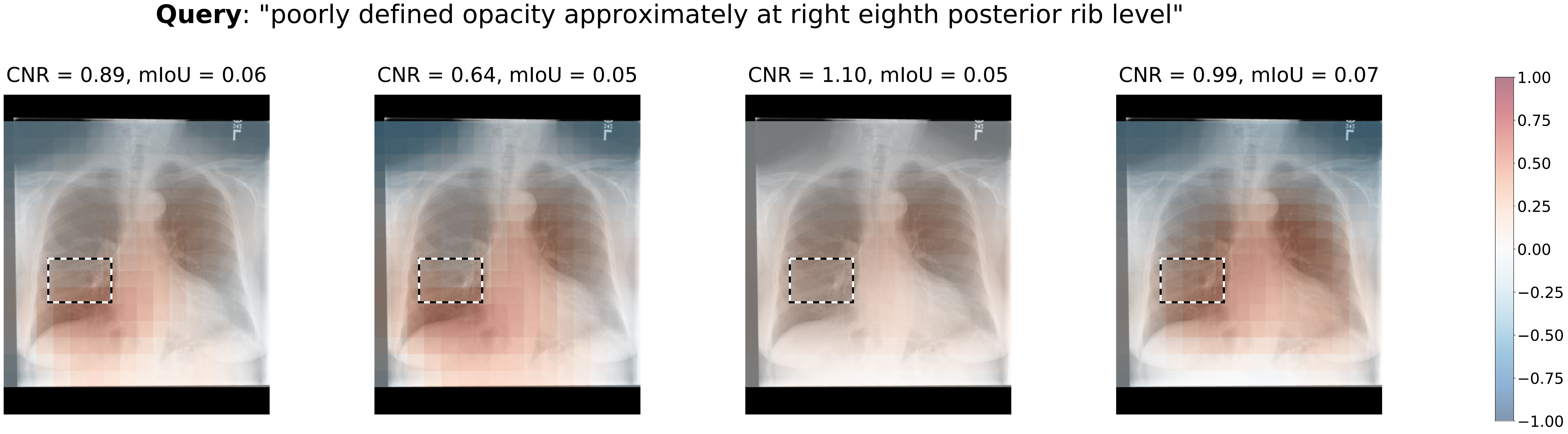}
        \caption{Failed to recognise the rib position}
    \end{subfigure} \\[1ex]
    \begin{subfigure}[b]{0.9\linewidth}
        \includegraphics[width=\textwidth]{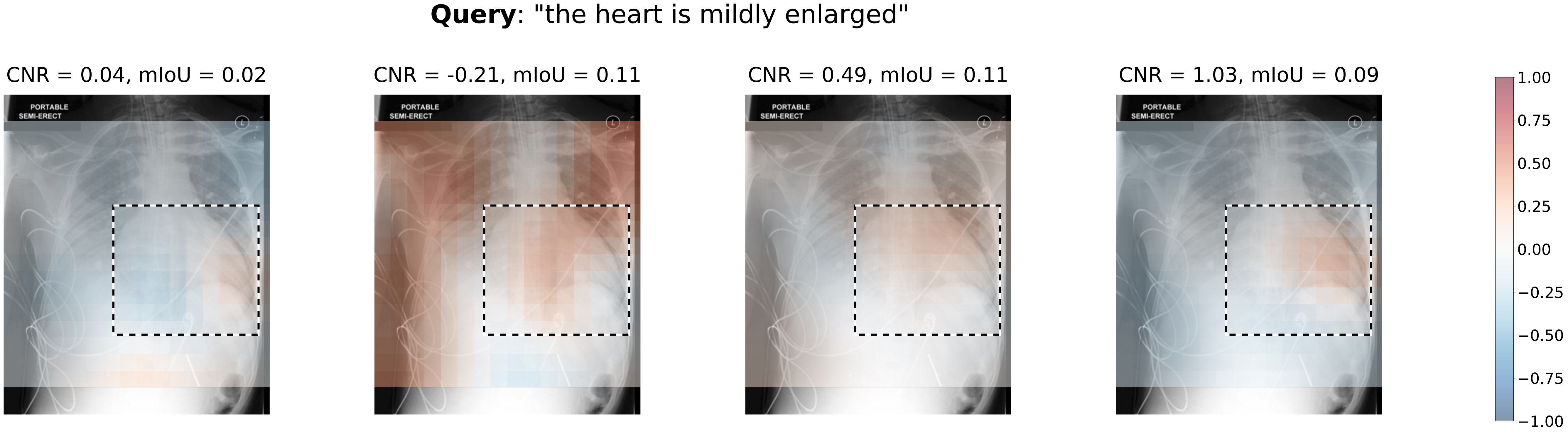}
        \caption{Mismatch between bounding box and salient region: Models attend to the salient region (enlarged area) to identify the abnormality instead of the entire heart.}
    \end{subfigure}
    \caption{Failure cases from \dataset~phrase grounding benchmark. Model outputs (latent  vector  similarity) are compared (from left, ClinicalBERT baseline, ConVIRT, GLoRIA, and \jointmodel)}
    \label{fig:failure_cases}
\end{figure}

Additionally, in \cref{fig:failure_cases}, we show and describe some failure cases of \jointmodel\ on the \dataset\ dataset to motivate any further research on this topic. In particular, the models show limitations in grounding the descriptions relating to smaller structures (e.g., rib fracture, pneumothorax), and in a few cases the location modifier is not disassociated from the entities corresponding to abnormalities, see (a) in \cref{fig:failure_cases}.

\begin{table}[t]
\centering
\caption{An extension of \cref{table:rsna_classification} to include Sensitivity and Specificity for the RSNA Pneumonia zero-shot and fine-tuned classification. We compare to GLoRIA scores reported in \cite{huang2021gloria}  which outperforms ConVIRT~\cite{zhang2020contrastive} (see ~\cite{huang2021gloria}). Training size: GLoRIA ($N=186k$, private dataset), \jointmodel\ ($N=146.7k$ of MIMIC-CXR).}\label{table:rsna_classification_full}
\vspace{3pt}
\centering
\setlength{\tabcolsep}{3pt}
\scriptsize
\resizebox{\linewidth}{!}{
\begin{tabular}{@{}llclcccccc@{}}
    \toprule
Method & Type & Text Model & Loss & \% of labels & Acc. & Sens. & Spec. & F1 & AUROC \\
\midrule
\multirow{3}{6em}{SimCLR~\cite{sicmlr}} & \multirow{3}{3.5em}{Image only} & \multirow{3}{*}{-} & \multirow{3}{*}{Global} & 1\% & 0.545 & 0.776 & 0.436 & 0.522 & 0.701 \\
 &&&& 10\% & 0.760 & 0.663 & 0.806 & 0.639 & 0.802 \\
 &&&& 100\% & 0.788 & 0.685 & 0.837 & 0.675 & 0.849 \\
 \midrule
\multirow{4}{6em}{GLoRIA~\cite{huang2021gloria}} & \multirow{4}{*}{Joint} & \multirow{4}{*}{ClinicalBERT}& \multirow{4}{3.5em}{Global \& local}
 & Zero-shot & 0.70 & 0.89 & 0.65 & 0.58 & - \\
 &&&& 1\% & 0.72 & 0.82 & 0.69 & 0.63 & 0.861 \\
 &&&& 10\% & 0.78 & 0.78 & 0.79 & 0.63 & 0.880 \\
 &&&& 100\% & 0.79 & 0.87 & 0.76 & 0.65 & 0.886 \\
 \midrule
 Baseline & Joint & ClinicalBERT & Global & Zero-shot & 0.719 & 0.648 & 0.781 & 0.614 & 0.812\\
 \midrule
\multirow{4}{*}{\jointmodel} & \multirow{4}{*}{Joint} & \multirow{4}{*}{CXR-BERT}  & \multirow{4}{*}{Global} & Zero-shot & 0.732 & 0.831 & 0.685 & 0.665 & 0.831  \\
 &&  &                  & 1\%       & 0.805 & 0.791 & 0.812 & 0.723 & 0.881 \\
 &&  &                  & 10\%      & 0.812 & 0.781 & 0.826 & 0.727 & 0.884 \\
 &&  &                  & 100\%     & 0.822 & 0.755 & 0.856 & 0.733 & 0.891 \\
\bottomrule
\end{tabular}}
\vspace{-10pt}
\end{table}

\subsection{Additional Evaluation Metrics}
In \cref{table:rsna_classification_full}, an extension of \cref{table:rsna_classification} is provided to include the sensitivity and specificity metrics for the zero-shot and fine-tuned classification experiments presented in \cref{sec:rsnaclassification}. The classification thresholds are set to maximise the F1 scores for each method. Further, in \cref{table:grounding_iou_minmax_rescaling} we provide mean IoU scores for the phrase grounding experiments presented in \cref{sec:phrase_grounding}, which evaluates the pretrained \jointmodel\ model on the \dataset\ dataset.
We observed that the distribution of similarity scores is different for GLoRIA and \localjointmodel\ due to the different temperature parameter used in the local loss term in  \cite{huang2021gloria}.
To provide a fair comparison, we adjust the similarity scores via min-max scaling to the full $[-1,1]$ range. The same scaling strategy is utilised in the implementation of the baseline method~\cite{huang2021gloria}. Note that the CNR scores are not affected by this linear re-scaling.

\subsection{Ablations on Training Dataset Size \& Use of Raw Input Images}

An additional set of experiments are conducted to test the impact of (I) training dataset size and (II) the use of raw DICOM images instead of JPEG images on phrase grounding performance. In the former case, the number of training pairs is increased from $146.7k$ to $176k$, where we used all available studies with IMPRESSION section and AP/PA scans after excluding the test set. In the latter ablation, the JPEG images are replaced with the raw DICOM images to reduce image artefacts due to compression. \Cref{table:phrasegrounding_ablations} shows that further performance gains can be achieved by utilising the DICOM data and matching the training set size to related methods (e.g., GLoRIA \cite{huang2021gloria}), where the raw data is empirically observed to contribute more. These improved results and pre-training models are neither reported nor used in the experiments presented in the main body of this paper. We hope that these findings can provide useful insights for future research on this topic.

\begin{table}[t]
\vspace{-3pt}
\caption{Ablations on \jointmodel\ -- Increasing training set size and use of raw DICOM images instead of compressed JPEG images. The approaches are compared in terms of contrast-to-noise ratio (CNR) obtained on the newly released \dataset\ dataset, averaged over four runs with different seeds.}
\resizebox{.99\textwidth}{!}{
\begin{tabular}{@{}lllccccccccc@{}}
    \toprule
 Method & Training & Atelectasis & Cardiomegaly & Consolidation & Lung opacity & Edema & Pneumonia & Pneumothorax & Pl. effusion & Avg.\\
\midrule
\jointmodel   & 146.7k &  1.02$\pm$.06 & 0.63$\pm$.08 & 1.42$\pm$.02 & 1.05$\pm$.06 & 0.93$\pm$.03 & 1.27$\pm$.04 & 0.48$\pm$.06 & 1.40$\pm$.06 & 1.03$\pm$.02\\
+ More data   & 176.0k &  1.01$\pm$.07 & 0.70$\pm$.03 & 1.45$\pm$.01 & 1.04$\pm$.04 & 0.94$\pm$.01 & 1.27$\pm$.05 & 0.54$\pm$.05 & 1.43$\pm$.04 & 1.05$\pm$.02\\
+ Raw images  & 176.0k &  1.03$\pm$.06 & 0.64$\pm$.09 & 1.51$\pm$.02 & 1.12$\pm$.06 & 1.00$\pm$.07 & 1.39$\pm$.04 & 0.56$\pm$.05 & 1.46$\pm$.05 & 1.09$\pm$.02\\
\bottomrule
\end{tabular}}{}
\label{table:phrasegrounding_ablations}
\end{table}

\begin{table}[tb]
\vspace{-6pt}
\centering
\caption{Mean IoU  scores obtained on the newly released \dataset\ dataset, averaged over four runs with different seeds. The results are collected using different text encoder and training objectives (e.g., G\&L: Global and local loss). 
}
\resizebox{.99\textwidth}{!}{
\begin{tabular}{@{}lllccccccccc@{}}
    \toprule
 Method & Objective & Text encoder & Atelectasis & Cardiomegaly & Consolidation & Lung opacity & Edema & Pneumonia & Pneumothorax & Pl. effusion & Avg.\\
\midrule
Baseline                            & Global & ClinicalBERT & 0.228 & 0.269 & 0.293 & 0.173 & 0.268 & 0.249 & 0.084 & 0.232 & 0.224 \\
Baseline                            & Global & PubMedBERT   & 0.225 & 0.293 & 0.297 & 0.167 & 0.266 & 0.286 & 0.077 & 0.222 & 0.225 \\
ConVIRT \cite{zhang2020contrastive} & Global & ClinicalBERT &  0.257 & 0.281 & 0.313 & 0.177 & 0.272 & 0.238 & 0.091 & 0.227 & 0.238 \\
GLoRIA \cite{huang2021gloria}       & G\&L & ClinicalBERT   &  0.261 & 0.273 & 0.324 & 0.198 & 0.251 & 0.246 & 0.100 & 0.254 & 0.246 \\
\midrule
\jointmodel                         & Global & \cxrmodel    &  0.296 & 0.292 & 0.338 & 0.202 & 0.281 & 0.323 & 0.109 & 0.290 & 0.266 \\
\localjointmodel                         & G\&L & \cxrmodel      &  0.302 & 0.375 & 0.346 & 0.209 & 0.275 & 0.315 & 0.135 & 0.315 & 0.284 \\

\bottomrule
\end{tabular}}{}
\label{table:grounding_iou_minmax_rescaling}
\end{table}

\section{Background in Chest Radiology}\label{appendix:cxrdata}
\vspace{-5pt}
Chest X-rays are the most commonly performed diagnostic X-ray examination, and a typical text report for such an exam consists of three sections: a ``Background'' section describing the reason for examination and the exam type, a ``Findings'' section describing abnormalities as well as normal clinical findings in the scan, and an ``Impression'' section which summarises the findings and offers interpretation with possible recommendations.
Multiple large Chest X-ray datasets have been released to the public (see \cite{tamliterati2020} for an overview of CXR image datasets), including multi-modal ones of images and text such as MIMIC-CXR~\cite{johnson2019mimic}, some also accompanied by small sets of expert-verified ground-truth annotations of various nature, making the application a popular candidate for exploring self-supervised VLP on biomedical data. 

The application area also possesses a strong clinical motivation. Globally, there is a shortage of qualified trained radiologists and a constantly increasing number of examinations in healthcare systems, workflows are hampered by issues such as a lack of standardisation in report writing, and fatigue-based errors occur too frequently. Thus, decision-support systems that can analyse incoming images or image-report pairs in order to provide real-time feedback to radiologists are a promising avenue towards improving workflow efficiency and the quality of medical image readings. In practice, the existing radiology workflow can for example be augmented via machine learning models by providing feedback on any incorrect or missing information in reports, and by standardising the reports' structure and terminology.

\subsection{Key NLP and Dataset Challenges in Radiology}\label{sec:background}
In this work, we focus on developing text and image models to enable clinical decision-support systems for biomedical applications via self-supervised VLP, without ground-truth annotations, and we conduct experiments in CXR applications. Image and text understanding in the biomedical domain is distinct from general-domain applications and requires careful consideration. Medical images are elaborately structured, which is reflected in the corresponding notes. To be able to harness the dense information captured in text notes for free-text natural language supervision, it becomes imperative to obtain finely tuned text models. 

\paragraph{Complex Sentence Structure.}
Linguistic characteristics in radiology reports, many shared with related clinical text settings, decidedly differ from general domain text and thus require carefully tuned text models to acquire the best possible free-text natural language supervision in self-supervised VLP.  For one, negations are frequently used to indicate the absence of findings, in particular to make references as to how a patient's health has evolved, e.g. ``there are no new areas of consolidation to suggest the presence of pneumonia''. This sentence is for example falsely captured as positive for pneumonia by the automated CheXpert labeller~\cite{irvin2019chexpert}.  
Furthermore, as exemplified in this example, long-range dependencies are common, which makes understanding of relations within sentences challenging. 

\paragraph{Use of Modifiers.}
Another characteristic is the use of highly specialised spatial language in radiology, which is crucial for correct diagnosis, often describing the positioning of radiographic findings or medical devices with respect to anatomical structures, see e.g. ~\cite{datta2020hybrid,datta2020understanding}. The use of words like ``medial'', ``apical'', ``bilateral'' or ``basilar'' as spatial modifiers is unlikely to appear in the general domain but very common in CXR radiology reports. In addition to spatial modifiers, severity modifiers such as ``mild'', ``moderate'' or ``severe'' are also commonly attached to an identified disorder or abnormality~\cite{dligach2014discovering}. 

\paragraph{Expressions of Uncertainty.}
Another interesting difference to most general domain VLP applications and datasets such as Internet image captions, are expressions of uncertainty that one frequently encounters in radiology reports. We rarely expect to find an image caption to read ``We see a person petting an animal, it is likely a dog but it could also be a cat''. In contrast, consider the following real radiology example: ``New abnormality in the right lower chest could be either consolidation in the lower lobe due to rapid pneumonia or collapse, and/or moderate right pleural effusion, more likely abnormality in the lung because of absent contralateral mediastinal shift.''  It is an extremely long description expressing uncertainty and containing long range dependencies. 

\paragraph{Class Imbalance.}
Finally, a challenge for many domain-specific VLP applications that is far less pronounced in the general domain setting is that of imbalanced latent entities. An example of such entities are the normal and anomalous findings in radiology images that doctors will describe in their report. In the CXR application, reports can roughly be divided into normal and abnormal scans, where abnormal ones reveal signs or findings observed during the exam~\cite{dai2021bdkg}. Normal scans that do not show any signs of disease are far more common than any other findings, which leads to a larger number of false negatives in contrastive objectives compared to the general domain. An important detail is that normal scans tend to be expressed in specific forms and doctors frequently use templates to produce reports with no abnormalities.

\section{MS-CXR Dataset Details}\label{appendix:ourdataset}

\begin{figure}[p]
    \centering
    \begin{subfigure}[b]{\linewidth}
        \includegraphics[width=\textwidth]{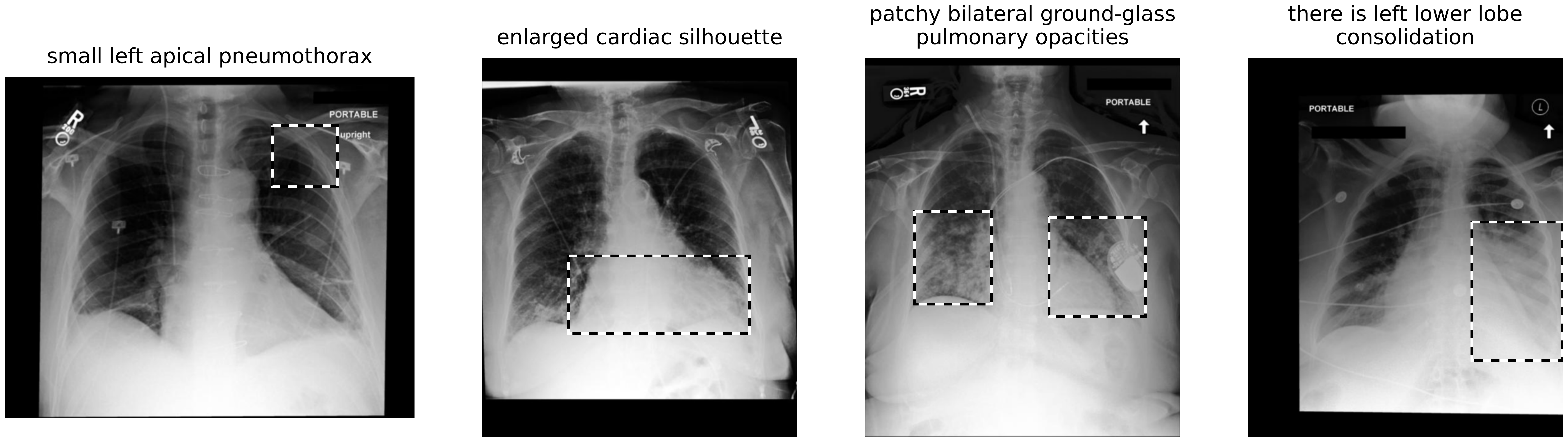}
        \caption{Spatial extent of abnormalities ranging from highly localised to large and diffuse}
    \end{subfigure} \\[2ex]
    \begin{subfigure}[b]{\linewidth}
        \includegraphics[width=\textwidth]{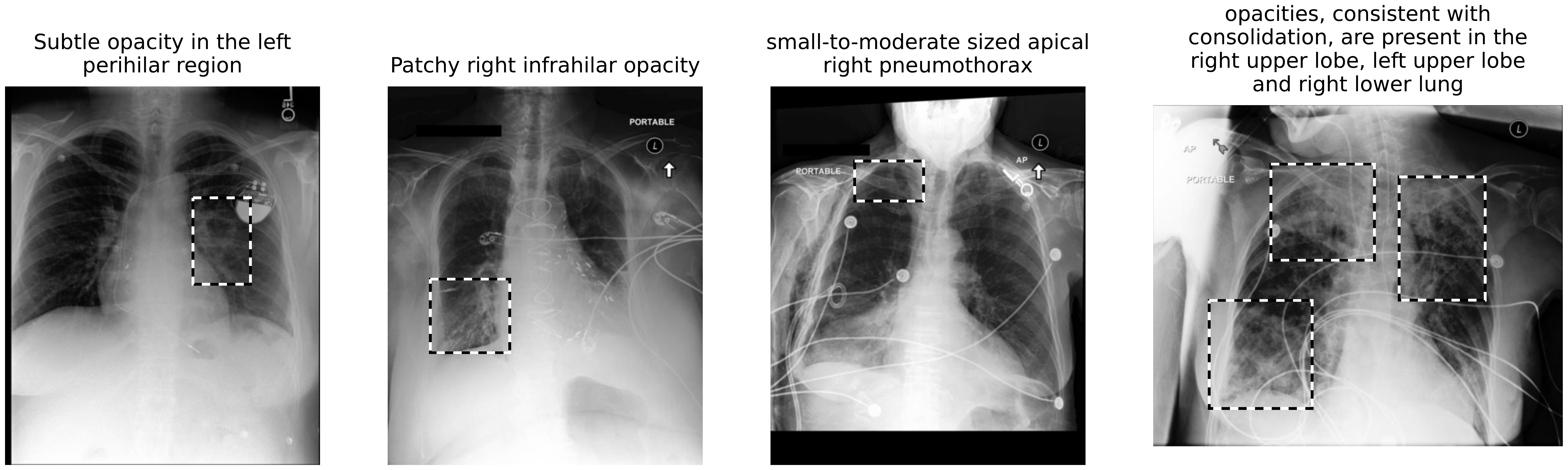}
        \caption{Complex spatial modifiers commonly seen in radiology reports}
    \end{subfigure} \\[2ex]
    \begin{subfigure}[b]{0.48\linewidth}
        \includegraphics[width=\textwidth]{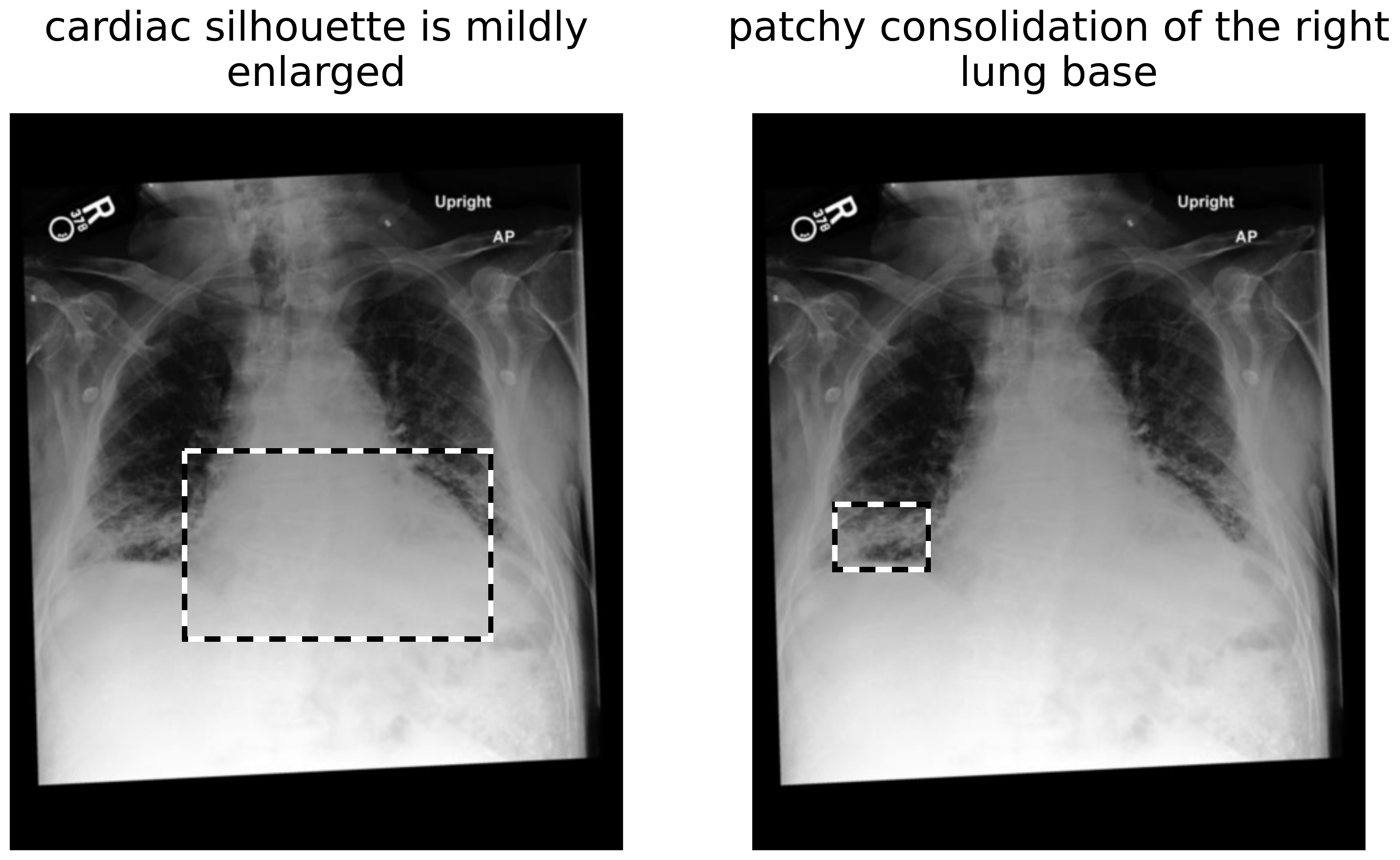}
        \caption{Multiple pathologies reported for the same study}
    \end{subfigure}
    \hfill
    \begin{subfigure}[b]{0.48\linewidth}
        \includegraphics[width=\textwidth]{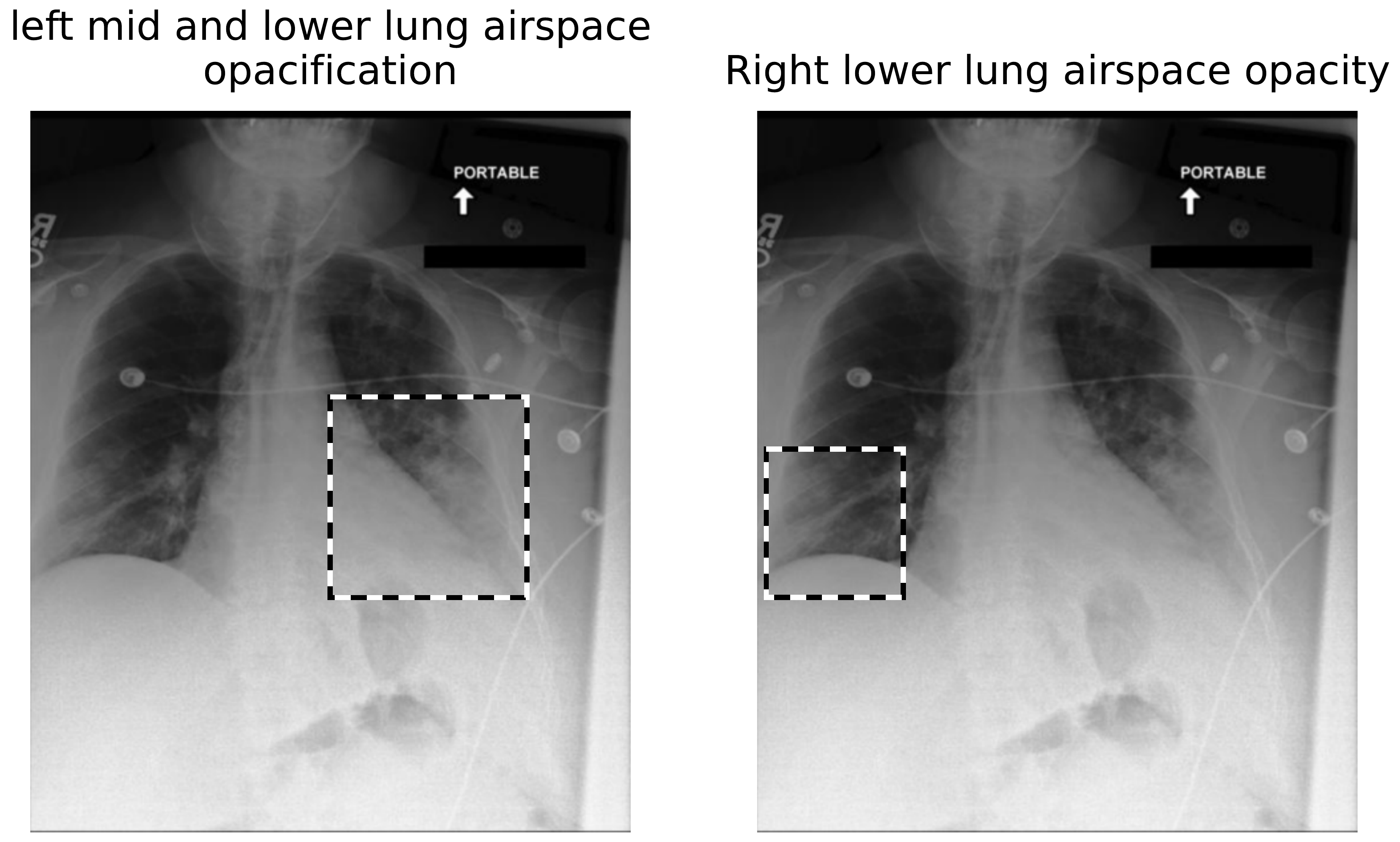}
        \caption{Findings with multiple spatial locations reported separately}
    \end{subfigure}
    \caption{We here provide some examples illustrating important axes of variability present in the \dataset\ dataset. Text descriptions include clinical findings of varying spatial extent (a) and a range of different spatial modifiers (b). Additionally, a subset of studies contain multiple bounding-box and sentence annotations per image (c--d).}
    \label{fig:datasetexamples}
\end{figure}

\paragraph{General Overview.} With this new benchmark dataset, we provide bounding box and sentence pair annotations describing clinical findings visible in a given chest X-ray image. \dataset\ consists of $1047$ images, with a total of $1153$ bounding box and sentence pairs.
Each sentence describes a single pathology present in the image, and there could be multiple manually annotated bounding boxes corresponding to the description of the single radiological finding. Additionally, an image may have more than one pathology present, and we provide separate sets of bounding boxes for each phrase describing a unique pathology associated with an image. The annotations were collected on a subset of MIMIC-CXR images, which additionally contains labels across eight different pathologies: atelectasis, cardiomegaly, consolidation, edema, lung opacity, pleural effusion, pneumonia and pneumothorax. These pathologies were chosen based on the overlap between pathology classes present in the existing datasets and the CheXbert classifier~\cite{smit2020combining}. In \cref{fig:datasetexamples} and \cref{tab:dataset_sentence_examples}, we show some representative image and text examples from \dataset. Additionally, the distribution of samples across the pathology classes is shown in \cref{table:benchmark_stats_demographics} together with demographics across subjects in \dataset. 

\paragraph{Differences to Existing Annotations.} The proposed benchmark builds on top of publicly available bounding-box/ellipse annotations in REFLACX~\cite{lanfredi2021reflacx} and MIMIC-CXR-Annotations~\cite{tamliterati2020}, where the latter also contains simplified text phrases for pneumonia and pneumothorax. \dataset\ extends and curates these annotation sets by (I) adding a new set of studies to cover a wider range of clinical findings and pathologies, (II) reviewing the clinical correctness and suitability of the existing annotations for the grounding task (see \Cref{sec:ourdataset}), 
(III) creating, verifying, and correcting bounding boxes where necessary, and (IV) pairing them up with real clinical descriptions extracted from MIMIC-CXR reports if none were present. Most importantly, the textual descriptions paired with dense image region annotations are sampled from the original distribution of word tokens, which capture dense text semantics and are better aligned with real-world clinical applications that build on good local alignment.

\begin{table}[t]
\centering
\caption{Distribution of the annotation pairs (image bounding-box and sentence) across different clinical findings. The demographic statistics (e.g., gender, age) of the subjects are collected from MIMIC-IV dataset for \dataset\ and all MIMIC-CXR.}
\resizebox{.8\textwidth}{!}{
\begin{tabular}{@{}lcccc@{}}
\toprule
Findings & \# of annotation pairs\; & \# of subjects & Gender - F (\%) &  Avg Age (std)\\
\midrule
Atelectasis & 61 & 61 & 28 (45.90\%) & 64.52 (15.95) \\
Cardiomegaly & 333 & 282  & 135 (47.87\%) & 68.10 (14.81) \\
Consolidation & 117 & 109 & 40 (36.70\%) & 60.08 (17.67) \\
Edema & 46 & 42 & 18 (42.86\%) & 68.79 (14.04) \\
Lung opacity &  81 & 81 & 33 (40.24\%) & 62.07 (17.20) \\
Pleural effusion & 96 & 95 & 41 (43.16\%) & 66.36 (15.29) \\
Pneumonia & 182 & 146  & 65 (44.52\%) & 64.32 (17.17) \\
Pneumothorax & 237 & 151 & 66 (43.71\%) & 60.71 (18.04) \\
\midrule
Total & 1153 &  851 & 382 (44.89\%) & 64.37 (16.61) \\
Background (all MIMIC-CXR) & - & 65379  & 34134.0 (52.39\%) & 56.85 (19.47) \\
\bottomrule
\end{tabular}}{}
\label{table:benchmark_stats_demographics}
\end{table}

\subsection{Label Collection and Review}\label{appendix:ourdataset_label_collection}
We first parse original MIMIC reports and REFLACX \cite{lanfredi2021reflacx} radiology transcripts by extracting sentences to form a large pool of text descriptions of pathologies. These candidates are later filtered by deploying the CheXbert~\cite{smit2020combining} text classifier, in order to keep only the phrases associated with the target pathologies whilst ensuring the following two criteria: (I) For a given study, there is only one sentence describing the target pathology, and (II) the sentence does not mention more than one findings that are irrelevant to each other. After extracting the text descriptions, they are paired with image annotations on a study level. At the final stage, a review process is conducted with two board certified radiologists mainly to verify the match between the text and  bounding box candidates. Moreover, in this review process, we also assessed the suitability of the annotation pairs for the grounding task whilst ensuring clinical accuracy. In detail, the phrase-image samples are filtered out if at least one of following conditions is met:
\begin{enumerate}
\item Text describing a finding not present in the image.
\item Phrase/sentence does not describe a clinical finding or describes multiple unrelated abnormalities that appear in different lung regions.
\item There is a mismatch between the bounding box and phrase, such as image annotations are placed incorrectly or do not capture the true extent of the abnormality.
\item High uncertainty is expressed regarding reported findings, e.g. ``there is questionable right lower lobe opacity''.
\item Chest X-ray is not suitable for assessment of the finding or has poor image quality.
\item Text contains differential diagnosis or longitudinal information that prohibits correct grounding via the single paired image.
\item Sentences longer than 30 tokens, which often contain patient meta-information that is not shared between the two modalities (e.g., de-identified tokens).
\end{enumerate}
Note that we only filter out phrases containing multiple findings, not images with multiple findings. For instance, if an image contains both pneumonia and atelectasis, with separate descriptions for each in the report, then we create two instances of phrase-bounding box pairs. Among those candidate annotations automatically extracted from radiology reports~\cite{johnson2019mimic} or dictated transcripts~\cite{lanfredi2021reflacx}, 222 of out 817 were rejected and not included in \dataset. Here the raw text data were first processed with an algorithm to extract caption candidates for the review process. The same review process is applied to adjudicate the annotation pairs released in \cite{tamliterati2020}, and 53 out of 367 pairs were rejected and not included in \dataset.

To further increase the size of our dataset, and to balance samples across classes, additional CXR studies are sampled at random, conditioned on the underrepresented pathologies. The following procedure is applied to create the pairs of image and text annotations for these selected studies: Text descriptions are extracted using the same methodology outlined above, using MIMIC-CXR and ImaGenome datasets \cite{wu2021chest}, where the latter provides sentence extracts from a subset of MIMIC-CXR dataset for clinical findings. However, differently from the initial step, the corresponding bounding box annotations (either one or more per sentence) are created from scratch by radiologists for the finding described in the text, and the same filtering as above is applied by the annotator to discard candidates if the image and/or sentence is found unsuitable for the grounding task.

\begin{table}[t]
    \centering
    \caption{Example findings in \dataset\ with complex syntactic structures. Please note how radiological sentences are most often not just a simple statement of the form ``[class1, class2, ...]'' that can be parsed with a simple bag-of-words approach, as in typical natural image captioning benchmarks (e.g., ``A couple getting married'' retrieved from Flickr30k \cite{plummer2015flickr30k}).
    }
    \label{tab:dataset_sentence_examples}
    \scriptsize
    \setlength{\tabcolsep}{.5em}
    \newcommand{\rowskip}{\rule{0pt}{3ex}}
    \begin{tabular}{@{} >{``}p{6cm}<{"} ll @{}}
        \toprule
        \multicolumn{1}{@{}l@{}}{Sentence} & Difficulty & Class \\
        \midrule
        Abnormal opacity in the basilar right hemithorax is likely atelectasis involving the right lower and middle lobes &Complex syntactic structure& Atelectasis \\\rowskip
        Multisegmental lower lobe opacities are present, consistent with areas of consolidated and atelectatic lung &Complex syntactic structure& Atelectasis \\\rowskip
        Parenchymal opacification in the mid and lower lung & Less common expression& Pneumonia \\\rowskip
        Air bronchograms extending from the left hilum throughout the left lung which has the appearance of infection &Complex location description& Pneumonia \\\rowskip
        Persistent focal bibasilar opacities, most consistent with infection &Domain-specific modifier& Pneumonia \\\rowskip
        Widespread infection, less severe on the left & Location partially specified& Pneumonia \\\rowskip
        Airspace consolidation in the right upper, right middle and lower lobes &Multiple locations& Pneumonia \\\rowskip
        Subsegmental-sized opacities are present in the bilateral infrahilar lungs &Domain specific modifiers& Lung opacity \\\rowskip
        There continues to be a diffuse bilateral predominantly interstitial abnormality in the lungs with more focal vague opacity in the left upper peripheral lung &Complex syntactic structure& Lung opacity \\\rowskip
        Left apical pneumothorax & Domain-specific modifier& Pneumothorax\\\rowskip
        Fluid level posteriorly, which represents a loculated hydropneumothorax &  Domain-specific language& Pneumothorax\\\rowskip
        Mild-to-moderate left pneumothorax
         & Severity modifier & Pneumothorax \\\rowskip     
        There is no pulmonary edema or pneumothorax, but small pleural effusions are still present
        & Negated disease entities & Pleural effusion \\\rowskip
        Pleural effusions are presumed but impossible to quantify, except say they are not large&Complex sentence structure& Pleural effusion \\
        \bottomrule
    \end{tabular}
\end{table}

\paragraph{Analysis of Average Phrase Length.} The average number of tokens (inc. full-stop) across all phrases is calculated for each benchmark dataset to better understand the characteristics of the dataset and domain. In that regard, the phrases released in \cite{tamliterati2020} has an average of $6.76$ tokens per sample and \dataset\ has an average of $7.49$ of tokens per sample. The auto-extracted radiology sentences from transcriptions \cite{lanfredi2021reflacx} whereas has an average of $8.49$ tokens per sample. We observe that relatively long sentences auto-extracted from transcripts \cite{lanfredi2021reflacx} were rejected more often in the review process as they often describe multiple clinical findings located in different image regions. This observation further emphasises the importance of review process of annotation pairs by the domain experts.

\paragraph{Patient Demographics.} As shown in \Cref{table:benchmark_stats_demographics}, the average age of subjects in \dataset\ is higher than the average for all subjects in MIMIC-CXR. We explain this observation with the fact that we do not sample studies from healthy subjects that do not display any anomalous findings and who are statistically likely to be younger. Similarly, we do not expect gender bias to be present due to our sampling as none of the pathologies we sample are gender-specific. Overall \dataset\  does not deviate far from the MIMIC-CXR distribution. 

\begin{table}
    \centering
    \caption{Example findings in ImaGenome which would make grounding of phrases difficult.
    }
    \label{tab:imagenome}
    \scriptsize
    \setlength{\tabcolsep}{.5em}
    \newcommand{\rowskip}{\rule{0pt}{3ex}}
    \begin{tabular}{@{} >{``}p{6cm}<{"} p{3cm} l @{}}
        \toprule
        \multicolumn{1}{@{}l@{}}{Sentence} & Difficulty & Annotated Finding \\
        \midrule
        Even though Mediastinal veins are more distended, previous pulmonary vascular congestion has improved slightly, but there is more peribronchial opacification and consolidation in both lower lobes which could be atelectasis or alternatively results of recent aspiration, possibly progressing to pneumonia.
        & Multiple findings, uncertainty, different sub-parts of lung
        & Pneumonia \\
        \midrule
        Moderate right pleural effusion and bilateral  heterogenous airplace opacities, concerning for pneumonia.
        &
        Multiple findings, differing laterality
        & Pneumonia \\
        \midrule
        It could be an early infection
        & Region unclear & Pneumonia \\
        \midrule
        There is also a new small left-sided pleural effusion. &
        Differential diagnosis, there could be another effusion
        & Effusion \\
        \bottomrule
    \end{tabular}
\end{table}

\section{Related Work}\label{sec:app_relatedwork}
Here we provide a more detailed overview of related work to complement the brief review provided in the main article. 
\paragraph{Joint Image-Text Representation Learning.}
A variety of self-supervised VLP approaches have been proposed towards jointly learning visual and textual representations of paired data without supervision, such as frameworks using contrastive objectives \cite{gupta2020contrastive,li2021supervision,radford2021learning}, approaches based on joint transformer architectures \cite{li2020unicoder,li2019visualbert,lu2019vilbert,su2019vl}, self-supervised VLP with word-region alignment and language grounding~\cite{chen2020uniter}, and text prediction tasks to learn image features~\cite{desai2021virtex}. 
For example, \cite{radford2021learning} use a contrastive loss over embeddings of text and image pairs to train a model on large data collected from the internet ($\sim$400M pairs) enabling zero-shot transfer of the model to downstream tasks.
Some of the proposed approaches utilise a single architecture, usually a transformer, to learn a representation, following encoders for the individual modalities \cite{chen2020uniter,li2019visualbert,su2019vl}. Another common theme is the use of use cross-modal attention mechanisms to improve the aggregation of image regions in convolutional architectures  \cite{akbari2019multi,datta2019align2ground,gupta2020contrastive}. 

A number of different objectives have been explored for representation learning in VLP, including the prediction of words in image captions~\cite{joulin2016learning}, predicting phrase n-grams~\cite{li2017learning}, predicting of entire captions~\cite{desai2021virtex}, \emph{global} contrastive objectives defined on the embeddings of the entire image and text instances~\cite{zhang2020contrastive}, and combinations of global and \emph{local} contrastive terms~\cite{huang2021gloria,muller2021joint}, where local means that objectives are defined over text fragments (words or phrases) and image regions.

A task closely related to instance representation learning in VLP is \textit{phrase grounding}, also known as visual grounding, phrase localisation, local alignment, or word--region alignment. The goal here is to connect natural language descriptions to local \textit{image regions}. 
In a supervised learning setting such as in~\cite{mao2016generation,mu2021disentangled}, this problem requires expensive manual annotation for region--phrase correspondence. 
Thus, settings for visual grounding have been explored in which cross-modal pairs are the only form of supervision that is available~\cite{chen2020uniter,datta2019align2ground,fang2015captions,gupta2020contrastive,liu2021relation,wang2020maf}, i.e. the supervision signal is the knowledge of which caption belongs to which image. This setting of paired images and text has also been referred to as weakly supervision. 
Much of the general domain prior work on phrase grounding relies on off-the-shelf object-detection networks~\cite{chen2020uniter,datta2019align2ground,gupta2020contrastive,wang2020maf,yu2020cross,zhang2020counterfactual} such as Faster R-CNN~\cite{ren2015faster} which are pretrained on large labelled datasets to extract region candidates from images. This considerably simplifies the problem of matching regions to phrases as the set of possible regions to match can be assumed to be known, a luxury that is often unavailable in domain specific contexts.
\paragraph{Biomedical VLP Representation Learning.}
Several studies~\cite{hsu2018unsupervised,huang2021gloria,liao2021multimodal,muller2021joint,zhang2020contrastive} have explored joint representation learning for paired image and text data in the medical domain. 
Contrastive VIsual Representation Learning from Text (ConVIRT)
~\cite{zhang2020contrastive} uses a contrastive learning formulation for instance-level representation learning from paired medical images and text. The authors uniformly sample sentences and maximise their similarity to true augmented paired images via the InfoNCE contrastive loss~\cite{oord2018representation}, while reducing similarity between negative pairs in the same batch.
\cite{huang2021gloria,muller2021joint} both introduce approaches that combine instance-level image--report contrastive learning with local contrastive learning for medical data. 
In contrast, \cite{liao2021multimodal} use a local-only objective in an approach that approximates the mutual information between grid-like local features of images and sentence-level text features of medical data. The formulation learns image and text encoders as well as a discriminator trained to distinguish positive and negative pairs.
While most related approaches use no ground truth, \cite{chauhan2020joint} study a semi-supervised edema severity classification setting,  and \cite{hayat2021multi} assume sets of seen and unseen labels towards zero-shot classification on CXR data.
\cite{li2020comparison} evaluate pretrained joint embedding models---general domain VLP representation learning models that use a transformer to learn a joint embedding---by fine-tuning the models on CXR data. 

Multiple CXR datasets exist that enable a partial evaluation of phrase grounding, but all come with some limitations we hope to mitigate with our \dataset\ dataset (see \cref{sec:ourdataset}). VinDr~\cite{nguyen2020vindr}, RSNA Pneumonia~\cite{shih2019augmenting}, and the NIH Chest X-ray Dataset~\cite{wang2017chestx} are datasets that provide bounding-box image annotations, but lack accompanying free-text descriptions. REFLACX~\cite{lanfredi2021reflacx} provides gaze locations captured with an eye tracker, dictated reports and some ground truth annotations for gaze locations, but no full phrase matches to image regions. Phrase annotations for MIMIC-CXR data released in~\cite{tamliterati2020} are of small size (350 studies), only contain two abnormalities, and for some samples have shortened phrases that were adapted to simplify the task. 
ImaGenome~\cite{wu2021chest} provides a large number of weak local labels for CXR images and reports, with a focus on anatomical regions. However, its ground-truth set is smaller (500 studies), bounding-box regions annotate anatomical regions rather than radiological findings. Furthermore, ImaGenome sentence annotations are not curated, see  \cref{tab:imagenome} for some examples. Sentences often contain multiple diseases as well as uncertain findings, making an accurate, largely noiseless grounding evaluation difficult. Some sentences also contain differential diagnosis and temporal change information, which cannot be grounded without access to prior scans.

\paragraph{Language Modelling in Radiology.} Most recent general domain VLP work relies on transformer based contextual word embedding models, in particular BERT~\cite{devlin2018bert}, pretrained on general domain data from newswire and web domains such as Wikipedia. 
But specific domains often exhibit differences in linguistic characteristics from general text and even related domains, such as between clinical and non-clinical biomedical text as noted in~\cite{alsentzer2019publicly}, motivating the use of more specialised language models in most related work with a focus on the medical domain. Here, related multi-modal work commonly uses publicly available models including BioBERT~\cite{lee2020biobert}, ClinicalBERT~\cite{alsentzer2019publicly},  BioClinicalBERT~\cite{alsentzer2019publicly}, or PubMedBERT~\cite{gu2021domain}, which are either trained from scratch or fine-tuned via continual pretraining using a Masked Language Modelling (MLM) objective. Sometimes additional objectives are added such as adversarial losses~\cite{liu2020adversarial} or Next Sentence Prediction. 
\cite{gu2021domain} provide evidence that training language models from scratch for specialised domains with abundant amounts of unlabelled text can result in substantial gains over continual pretraining of models first fit to general domain text. 
The specialised corpora these biomedical and clinical domain models use include PubMed abstracts and PubMed Central full texts, and de-identified clinical notes from MIMIC-III~\cite{johnson2016mimic}. 
All the aforementioned language models have a pre-specified vocabulary size consisting of words and subwords, usually 30,000 words in standard BERT. The in-domain vocabulary plays a particularly important role in representative power for a specialised domain. A vocabulary that is not adapted will break up more words into subwords and additionally contain word pieces that have no specific relevance in the specialised domain, hindering downstream learning (see e.g.~\cite{gu2021domain}). As \cite{gu2021domain} highlight, BERT models that use continual pretraining are stuck with the original vocabulary from the general-domain corpora. 

Other closely related tasks in the CXR domain that share similar NLP challenges include report summarisation~\cite{dai2021bdkg,zhang2018learning}, automatic report generation~\cite{chen2020generating,liu2019clinically,miura2021improving}, and natural language inference for radiology reports~\cite{miura2021improving}. Finally, while the name implies close similarity to our \cxrmodel, CheXbert~\cite{smit2020combining} is a BERT based sentence classification model developed for improving the CheXpert~\cite{irvin2019chexpert} labeller, and the model does not have a domain-specific vocabulary like ours or PubMedBERT. 

We note that most related work on self-supervised multi-modal learning on CXR data neither explores text augmentation nor maintains text losses such as MLM during multi-modal training. An exception is found in~\cite{muller2021joint}, who use the Findings and Impression/Assessment sections of radiology reports, and randomly change the sentence order by swapping pairs of them.

\section{Model Details}
\subsection{\cxrmodel \ Pretraining Details}
Our \cxrmodel~text encoder is based on the BERT (base size) architecture~\cite{vaswani2017attention}. We adopt an implementation available via the Huggingface transformers library~\cite{wolf2019huggingface}. The model weights are randomly initialised and pretrained from scratch. As described in \cref{sec:cxrmodel}, CXR-BERT is pretrained in three phases before the joint pretraining phase. For Phase (I), we use the Huggingface tokeniser library\footnote{\url{https://github.com/huggingface/tokenizers}} to generate our custom WordPiece vocabulary of 30k tokens. For Phase (II), we use the AdamW~\cite{loshchilov2018decoupled} optimiser with a batch size of 2048 sequences and a linear learning rate schedule over 250k training steps with a 5\% warm up period. We set a base learning rate of 4e-4. Following RoBERTa~\cite{liu2019roberta}, we pack multiple sentences into one input sequence of up to 512 tokens and use dynamic whole-word masking. In Phase (III), we continue pretraining the model using only MIMIC-CXR text reports. In addition to the MLM loss, we add our RSM loss to pretrain the projection layer. The projection layer $\projtxt$ is used to project the 768-dimensional feature vector $\precls$ to a 128-dimensional report representation $\cls$. We use the AdamW optimiser with a batch size of 256 sequences and a linear learning rate schedule over 100 epochs with a 3\% warm up period. We set the base learning rate to 2e-5.

\subsection{Image Encoder}

\paragraph{Pretraining Details.}
For the image encoder, we adopt the ResNet50~\cite{he2016deep} architecture. The 2048-dimensional feature maps $\localemb$ of the ResNet50 are projected to 128-dimensional feature maps $\localrep$ using a two-layer perceptron $\projimg$ implemented with $1 \times 1$ convolutional layers and batch-normalisation \cite{ioffe2015batch}. The global image representation $\globalrep$ is obtained by average-pooling the projected local features $\localrep$. Prior to image-text joint training, 
the model weights are randomly initialised and pretrained on MIMIC-CXR images using SimCLR~\cite{sicmlr} --- an image-only self-supervised learning approach. We use a large-batch optimisation (LARS) technique~\cite{you2017large} on top of ADAM with a batch size of 256 and a linear learning rate scheduler over 100 epochs with a 3\% warm up period. We set the base learning rate to 1e-3.

\begin{table}[t]
\centering
\caption{Hyper-parameter values used for image data augmentations.}
\setlength{\tabcolsep}{.5em}
\resizebox{.9\textwidth}{!}{
\begin{tabular}{@{}lccc@{}}
\toprule
 & Image-Text Pretraining & Image-only Pretraining & Fine-tuning for Downstream Tasks \\
\midrule
Affine transform -- shear & 15\textdegree & 40\textdegree & 25\textdegree \\
Affine transform -- angle & 30\textdegree & 180\textdegree & 45\textdegree \\
Colour jitter -- brightness & 0.2 & 0.2 & 0.2 \\
Colour jitter -- contrast & 0.2 & 0.2 & 0.2 \\
Horizontal flip probability & - & 0.5 & 0.5 \\
Random crop scale & - & (0.75, 1.0) & - \\
Occlusion scale & - & (0.15, 0.4) & - \\
Occlusion ratio & - & (0.33, 0.3) & - \\
Elastic transform ($\sigma, \alpha$)~\cite{simard2003best} & - & (4, 34) & - \\
Elastic transform probability & - & 0.4 & - \\
Gaussian noise & - & 0.05 & - \\
\bottomrule
\end{tabular}}{}
\label{table:imgaug_params}
\end{table}

\paragraph{Augmentations.}
For each training stage, we apply a different set of image augmentations to have a better control over the learnt feature invariances (e.g., laterality). During the image-text joint pretraining stage, we use affine transformations (random rotation and shearing) and contrast and brightness colour jitter. Unlike ConVIRT~\cite{zhang2020contrastive} and GLoRIA~\cite{huang2021gloria}, we do not apply horizontal flips during the joint training to preserve location information (e.g. ``pneumonia in the left lung''). During the image-only SSL (SimCLR) pretraining phase, we use additional image augmentations including random occlusion, additive Gaussian noise, and elastic spatial transforms~\cite{simard2003best}. We use the implementations available in the torchvision library\footnote{\url{https://pytorch.org/vision/stable/transforms.html}}. The image augmentation parameters and their corresponding values are listed in \cref{table:imgaug_params}. Before applying these transformations, we normalise the input image intensities by re-scaling each colour channel values to the $[0, 255]$ range. During inference, we only apply centre cropping and resizing.

\end{document}